%% file: sample.tex
\PassOptionsToPackage{hyperfootnotes=false}{hyperref}
\documentclass[twoside,11pt]{article}

\usepackage{blindtext}
\usepackage{wrapfig}
\usepackage{booktabs}

\usepackage{longtable} 
\usepackage[utf8]{inputenc}
\usepackage{CJKutf8}
\usepackage{listings}
\usepackage{minted}
\usepackage{geometry}
\usepackage[numbers]{natbib}
\usepackage{booktabs}
\usepackage{graphicx}
\usepackage{multirow}
\usepackage{subcaption}
\usepackage[protrusion=true,expansion=false]{microtype}

\usepackage[most]{tcolorbox}
\usepackage[scaled=1]{helvet}
\usepackage{lipsum} 

\tcbset{
    promptstyle/.style={
        enhanced,
        colback=white,
        colframe=black,
        colbacktitle=gray!20,
        coltitle=black,
        rounded corners,
        sharp corners=north,
        boxrule=0.5pt,
        drop shadow=black!50!white,
        attach boxed title to top left={
            xshift=-2mm,
            yshift=-2mm
        },
        boxed title style={
            rounded corners,
            size=small,
            colback=gray!20
        }
    },
    replystyleg/.style={
        enhanced,
        colback=green!15,
        colframe=black,
        colbacktitle=green!30,
        coltitle=black,
        boxrule=0.5pt,
        drop shadow=black!50!white,
        rounded corners,
        sharp corners=north,
        attach boxed title to top right={
            xshift=-2mm,
            yshift=-2mm
        },
        boxed title style={
            rounded corners,
            size=small,
            colback=green!40
        }
    },
    replystyler/.style={
        enhanced,
        colback=red!15,
        colframe=black,
        colbacktitle=red!40,
        coltitle=black,
        boxrule=0.5pt,
        drop shadow=black!50!white,
        rounded corners,
        sharp corners=north,
        attach boxed title to top right={
            xshift=-2mm,
            yshift=-2mm
        },
        boxed title style={
            rounded corners,
            size=small,
            colback=red!40
        }
    }
}
\newtcolorbox{promptbox}[1][]{
    promptstyle,
    title=Prompt,
    #1
}

\usepackage{dmlr2e}

\DeclareFontShape{OT1}{cmr}{bx}{sc}{%
  <5><6><7><8><9><10><10.95><12>%
  <14.4><17.28><20.74><24.88>cmcsc10%
}{}
\DeclareFontShape{OT1}{cmtt}{bx}{n}{%
  <5><6><7><8>cmtt8<9>cmtt9%
  <10><10.95>cmtt10%
  <12><14.4><17.28><20.74><24.88>cmtt12%
}{}
\newcommand{\dataname}[0]{\textsc{VisualSphinx}}

\usepackage{lastpage}
\dmlrheading{25}{2026}{1-\pageref{LastPage}}{11/25; Revised 08/26}{10/26}{21-0000}{Yichen Feng, Zhangchen Xu, Fengqing Jiang, Yuetai Li, Bhaskar Ramasubramanian, Luyao Niu, Bill Yuchen Lin, Basel Alomair, and Radha Poovendran} 
\def\openreview{\url{https://openreview.net/forum?id=3FHzSgBR4C}} 

\ShortHeadings{VisualSphinx: Large-Scale Synthetic Vision Logic Puzzles for RL
}{Feng et al.}
\firstpageno{1}

\begin{document}

\title{\dataname{}: Large-Scale Synthetic\\ Vision Logic Puzzles for RL}

\newcommand{\github}{\raisebox{-1.5pt}{\includegraphics[height=1em]{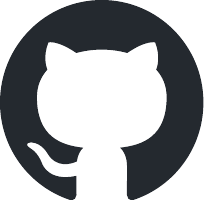}}}
\newcommand{\huggingface}{\raisebox{-1.5pt}{\includegraphics[height=1em]{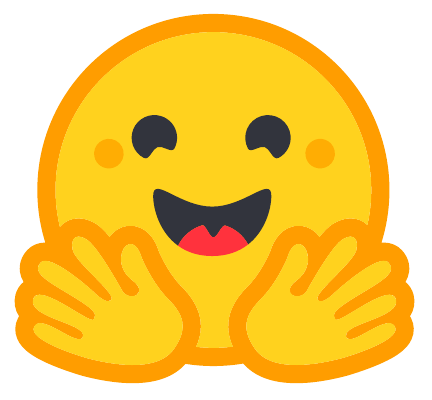}}}

\author{\name Yichen Feng\thanks{These authors contributed equally to this work.} \email yfeng42@uw.edu \\
        \addr University of Washington
        \AND
        \name Zhangchen Xu\footnotemark[1] \email zxu9@uw.edu \\
        \addr University of Washington
        \AND
        \name Fengqing Jiang \email fqjiang@uw.edu \\
        \addr University of Washington
        \AND
        \name Yuetai Li \email yuetaili@uw.edu \\
        \addr University of Washington
        \AND
        \name Bhaskar Ramasubramanian \email ramasub@wwu.edu \\
        \addr Western Washington University
        \AND
        \name Luyao Niu \email luyaoniu@uw.edu \\
        \addr University of Washington
        \AND
        \name Bill Yuchen Lin \email byuchen@uw.edu \\
        \addr University of Washington
        \AND
        \name Basel Alomair \email alomair@uw.edu \\
        \addr King Abdulaziz City for Science and Technology\\
        University of Washington\\
        HUMAIN
        \AND
        \name Radha Poovendran \email rp3@uw.edu \\
        \addr University of Washington\\
        \AND
        {\small{\github{} {\texttt{\url{https://visualsphinx.github.io}}}}}
        \quad {\small{\huggingface{} {\texttt{\url{https://hf.co/VisualSphinx}}}}}}

\editor{Yi Liu}

\maketitle

\input{sections/00-abstract}

\input{sections/01-intro}
\input{sections/02-related-work}
\input{sections/03-method}
\input{sections/04-analysis}
\input{sections/05-exp}
\input{sections/07-conclusion}

\impact{

\textbf{Positive impacts.}
\dataname{} offers a scalable, rule-grounded pipeline: structured rule abstraction, rule-level genetic diversification, programmatic rendering, and final assembly, yielding over \textbf{660K} visual logic puzzles at a total generation cost under \$1000. GRPO fine-tuning of \textsc{Qwen2.5-VL-7B} on \dataname{} improves puzzle-solving accuracy and shows positive but benchmark-dependent transfer to algebraic, arithmetic, and geometric tasks, helping democratize and systematize research on multimodal reasoning.

\textbf{Limitations.} 
The mechanisms underlying the observed improvements from RL training on \dataname{} remain underexplored. Our work primarily focuses on the generation and application of a large-scale synthetic dataset to improve VLM performance on visual reasoning. The experiments use single training runs, and the RL--SFT comparison does not match token exposure or compute budgets. Additionally, the current scope of \dataname~is limited to predefined categories of logical reasoning from seed questions, such as inductive, deductive, spatial, and structural logic. While these categories cover a broad range of visual reasoning tasks, they may not fully capture the complexity of real-world multi-modal reasoning scenarios.

\textbf{Ethical Considerations.} 
We have carefully reviewed the development and potential applications of \dataname and have not identified any specific ethical issues associated with this work.
}


\acks{
This work is partially supported by the National Science Foundation (NSF) AI Institute for Agent-based Cyber Threat Intelligence and Operation (ACTION) under grant IIS 2229876, the Air Force Office of Scientific Research (AFOSR) under grant FA9550-23-1-0208, and the Office of Naval Research (ONR) under grant N00014-23-1-2386.

This work is supported in part by funds provided by the National Science Foundation, Department of Homeland Security, and IBM. Any opinions, findings, and conclusions or recommendations expressed in this material are those of the author(s) and do not necessarily reflect the views of the NSF or its federal agency and industry partners.

Results presented in this paper were partially obtained using the Chameleon testbed \citep{keahey2020lessons} supported by the National Science Foundation.}

\vskip 0.2in
\bibliography{sample}

\clearpage
\appendix
\input{sections/appendix}

\end{document}

%% file: sections/00-abstract.tex
\vspace{-1em}
\begin{abstract}
    Vision language models (VLMs) are expected to perform effective multimodal reasoning and make logically coherent decisions, which is critical to tasks such as diagram understanding and spatial problem solving. 
    However, current VLM reasoning lacks large-scale and well-structured training datasets.
    To bridge this gap, we propose \dataname{}, a first-of-its-kind large-scale synthetic visual logical reasoning training dataset. To tackle the challenge of image synthesis with grounded answers, we propose a rule-to-image synthesis pipeline that extracts and expands puzzle rules from seed questions and generates executable rendering programs for puzzle assembly. Using standard GRPO with binary correctness rewards as a validation recipe, experiments show that VLMs trained on \dataname{} benefit from the logical coherence and readability of our dataset and exhibit improved performance on logical reasoning tasks.
    The enhanced reasoning capabilities developed from \dataname{} also benefit other reasoning tasks such as algebraic reasoning, arithmetic reasoning and geometry reasoning.
\end{abstract}

\begin{keywords}
  Vision language model, Logical reasoning, Synthetic data
\end{keywords}

\begin{figure}[t]
    \vspace{-0.5em}
    \centering
    \includegraphics[width=0.82\textwidth]{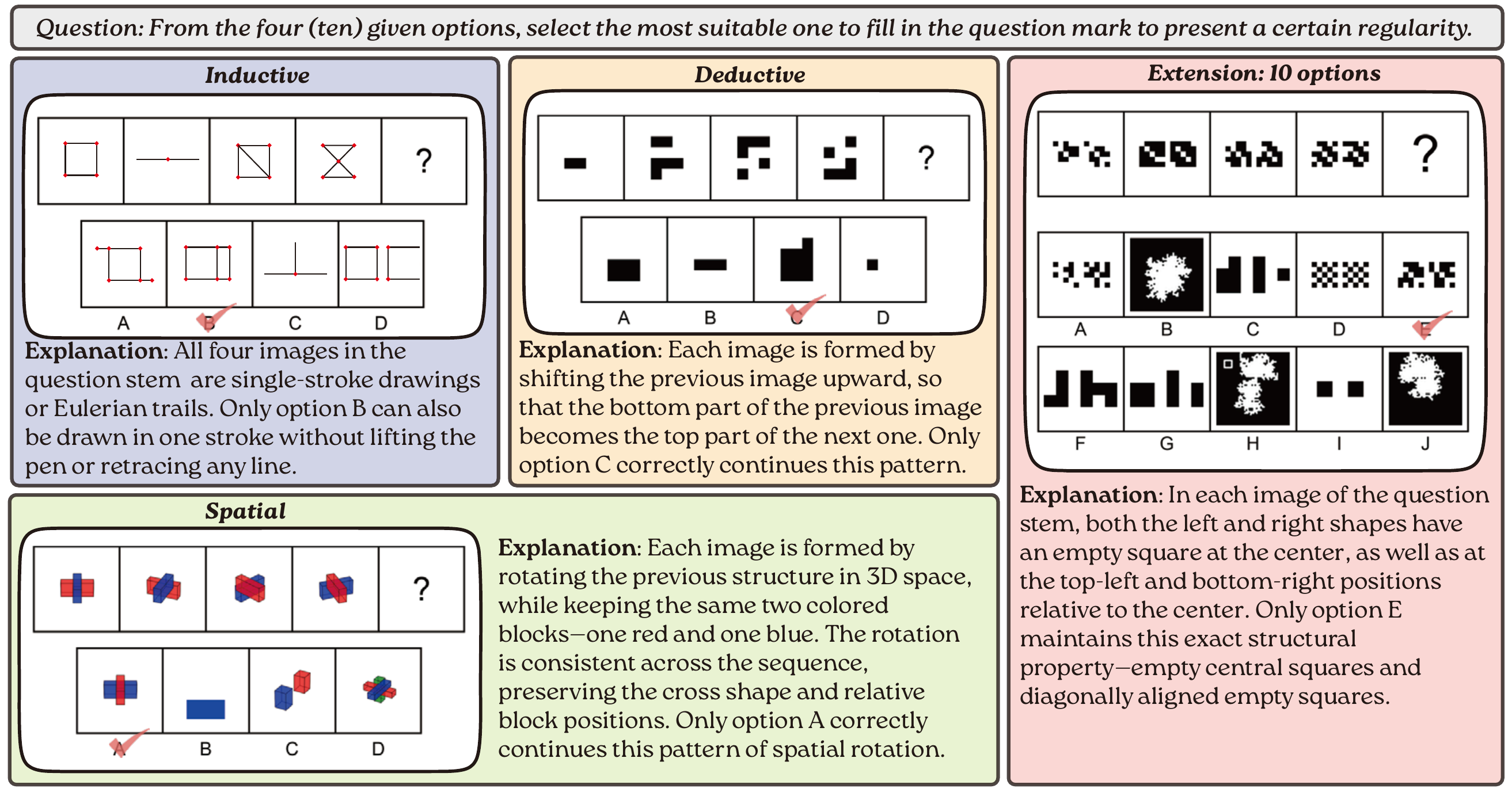}
    \caption{Example \dataname{} instances within each reasoning category. Each visual logic puzzle comprises a text prompt, a graphical question stem with four images and a question mark, and four candidate choices of graphical answers.}
    \label{fig:example}
    \vspace{-1.5em}
\end{figure}

%% file: sections/01-intro.tex
\section{Introduction}

Vision language models (VLMs) such as CLIP \citep{radford2021learning} and GPT-4V \citep{textocr-gpt4v, chen2024sharegpt4v} have demonstrated impressive capabilities in perception and generation tasks across modalities.
These models are capable of generating image captions \citep{olmo20252olmo2furious, li2024core}, answering complex questions about images \citep{openai2025o3o4mini}, and creating images based on textual descriptions \citep{openai2025introducing4o}.
Despite these advances, their reasoning abilities for tasks requiring structured logical inference over visual inputs remain limited~\citep{song2025visualpuzzles}. This shortcoming restricts their effectiveness in applications like diagram understanding \citep{li2025chain} and multimodal decision-making \citep{chen2023towards}, which demand the abstraction and generalization of visual patterns through coherent logic \citep{deng2024can, wang2025vision}.

Recently, reinforcement learning (RL) has been shown to be effective to enhance large language models' reasoning capabilities \citep{chen2025towards, zhou2025reinforced}.
However, RL is not readily applicable for VLMs due to the lack of large-scale training datasets with verified ground-truth responses.
Existing datasets such as Geo3K \citep{zheng2025easyr1} tailored for VLMs are often small in scale, require humans for curation, and lack structured annotations aligned with ground-truth.

In this paper, we bridge the gap by proposing \dataname{}, a large-scale synthetic dataset designed to train and benchmark VLMs for logical reasoning tasks.
\dataname{} consists of over \textbf{660K} automatically generated logical visual puzzles.
Each logical puzzle is grounded with an interpretable rule and accompanied by both correct answers and plausible distractors. To create this dataset, we propose a rule-level genetic algorithm to ensure diverse puzzle types, spanning inductive, deductive, spatial, and structural logic, followed by program-based image synthesis with varied styles to enhance visual diversity. Answer labels are determined during puzzle construction, while quality is assessed through VLM-based checks, deduplication, and removal of low-quality entries. Notably, the entire dataset is generated at a cost of less than \$1000, demonstrating exceptional scalability and cost-efficiency. Figure \ref{fig:example} provides examples covering diverse reasoning categories in \dataname{}. More examples and statistics are listed in Appendix~\ref{appendix: statistics}.

We perform GRPO on \textsc{Qwen2.5-VL-7B} model \citep{qwen2.5-VL} with \dataname{}. 
Experimental results demonstrate that the fine-tuned model substantially improves its accuracy on \dataname{}-\textsc{Bench} and exhibits positive but benchmark-dependent transfer in Section~\ref{section: experiments results}, including external visual logic, algebraic, arithmetic, and geometric reasoning tasks.
Our contribution is data-centric rather than algorithmic: we do not propose a new RL method. Instead, we use a standard GRPO setup with binary correctness rewards to test whether program-grounded and answer-verifiable visual logic puzzles provide useful RL signal for VLM reasoning.

This paper is structured as follows. Section~\ref{sec: related} reviews related work on logical reasoning and synthetic dataset generation. Section~\ref{sec: pipeline} describes the \dataname{} data curation pipeline. Section~\ref{sec: dataset statistics and analysis} presents a statistical analysis of the generated dataset’s quality. Section~\ref{sec: performance analysis} evaluates the performance of model trained on \dataname{}, demonstrating its effectiveness for enhancing VLM reasoning.
We conclude our paper in Section \ref{sec: conclusion}.

\begin{figure}[!tb]
    \centering
    \includegraphics[width=\textwidth]{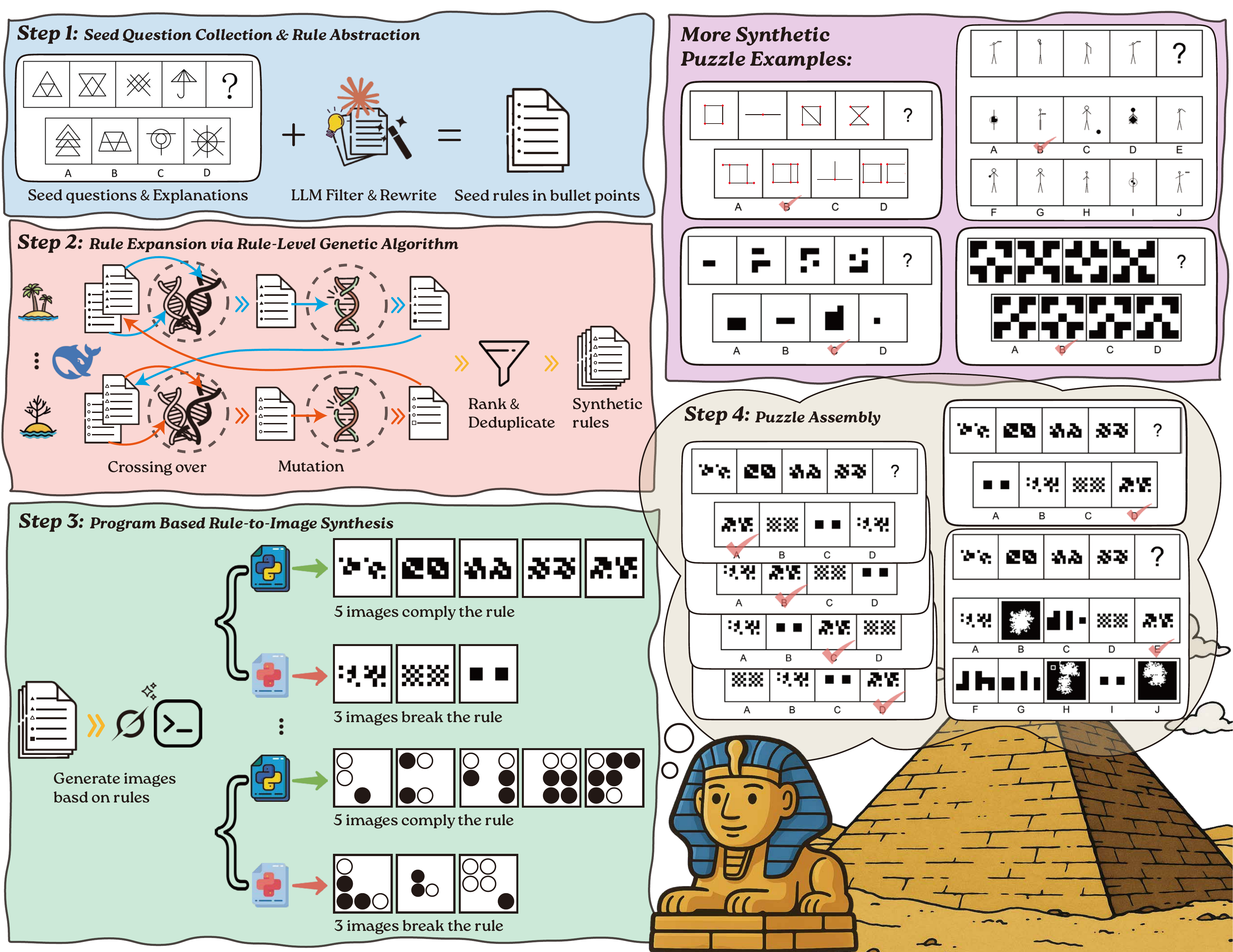}
    \vspace{0.5em}
    \caption{This figure illustrates the \textbf{four-stage} pipeline for generating \dataname{}. In Step 1, we collect 4K seed puzzles with explanations and abstract them into \textbf{structured rule} descriptions using LLMs. In Step 2, we apply a rule-level genetic algorithm to cross over, mutate and diversify the seed rules, scaling them to 40K high-quality rules. In Step 3, each rule is paired with a rendering style and used to generate five correct and three incorrect images via LLM-generated Python scripts. The fifth correct image is designated as the answer option, while the three rule-breaking images serve as distractors. After deduplication, we obtain 110K image groups. In Step 4, we assemble puzzles from each group using three complementary strategies: default assembly, shuffled answer variants, and expanded distractor sets. This results in over 660K visual logic puzzles, enabling robust and diverse training for multimodal reasoning models.}
    \vspace{-2em}
    \label{fig:main}
\end{figure}

%% file: sections/02-related-work.tex
\section{Related Work}
\label{sec: related}

\paragraph{Logical Reasoning.} Logical reasoning is a fundamental capability for large language models (LLMs). While existing research in math and coding seeks to measure this skill, such tasks often require substantial domain knowledge, making it challenging to isolate reasoning from subject expertise~\citep{cobbe2021training, chen2021evaluating, zhang2025unified, li2025star}. To address this, several studies focus on minimizing background knowledge demands. For instance, \citet{liu2020logiqa} employs multiple-choice logical comprehension questions, and \citet{lin-etal-2021-riddlesense} reformulates classic riddles into commonsense logic problems. Other works explore more structured puzzles, extending to programming challenges~\citep{schuster2021programming} and games such as \emph{Game of 24}~\citep{yao2023tree}, \emph{chess}~\citep{feng2023chessgpt}, and \emph{crosswords}~\citep{rozner2021decrypting}. Multimodal benchmarks also include logical puzzles; for example, \citet{lu2023mathvista} includes visual logic problems, while recent studies~\citep{ren2025vgrp, xu2025visulogic, zhang2025puzzlebench} extensively probe the ability of vision-language models (VLMs) to solve multimodal logic puzzles. Collectively, these resources isolate inference ability while reducing reliance on specialized knowledge.

\paragraph{Post-Training Dataset for VLMs.}  
Pretrained vision-language models (VLMs) require post-training to acquire the capability to solve complex tasks. Prior work has constructed large-scale instruction-tuning datasets, such as those by \citet{llava, ye-etal-2023-ureader}, to teach models to follow human instructions effectively. Subsequently, more targeted reinforcement learning (RL) is employed to further align VLMs with human feedback and enhance their long chain-of-thought reasoning capabilities~\citep{yang2025r1, MMEureka, zhang2025mm}. In contrast to these efforts, our \dataname{} dataset provides a large-scale multimodal training resource tailored specifically for visual logic puzzle reasoning.

\paragraph{Synthetic Visual RL Data.}
Recent work has begun to study rule-based RL for VLMs using fully synthetic, verifiable visual tasks. \citet{wang2025jigsaw} study visual RL with jigsaw puzzles, where task rules provide inherent ground truth and support controlled generalization analysis. \citet{alam2025sphinx} introduce a procedural visual perception and reasoning environment with verifiable solutions across task families such as symmetry, transformation, spatial reasoning, chart interpretation, and sequence prediction. \dataname{} shares the goal of providing verifiable visual RL signal, but focuses on exam-style visual logic puzzles. Its pipeline expands explicit seed rules at the semantic-rule level, renders rule-following answers and rule-violating distractors through executable programs, and supports both RL training and held-out evaluation within a controlled visual-logic distribution. 

\paragraph{Dataset Synthesis for VLM Training.}
Dataset synthesis has been widely used to expand LLM training resources by distilling knowledge from strong LLMs or refining their generated data \citep{wang-etal-2023-self-instruct,xu2023wizardlm,xu2024magpie}.
To close the gap in high-quality image–text instruction datasets, prior works \citep{fang2024vila, chen2024allava,  luo2024mmevol, zhang2024multimodal} incorporate human annotators or strong VLMs into synthesis pipelines, thereby enhancing the quality of large-scale synthetic datasets for VLM training.

%% file: sections/03-method.tex
\section{\dataname{}: Synthesizing Large-scale Vision Logic Post-Training Data for Visual Reasoning}
\label{sec: pipeline}

\textbf{Overview of \dataname{} Pipeline.} In what follows, we describe our scalable and efficient visual data generation method to synthesize visual logic puzzles with correct answers for RL training. Each visual logic puzzle consistently comprises a text prompt, a graphical question stem with four images and a question mark, and four candidate choices of graphical answers. As shown in Figure \ref{fig:main}, the pipeline consists of four steps: (1) seed question collection \& rule abstraction, (2) rule expansion via rule-level genetic algorithm, (3) program based rule‑to‑image synthesis and (4) puzzle assembly. The \dataname{} pipeline is fully automated and generates diverse visual logic puzzles across multiple pattern categories for multimodal RL training.

\subsection*{Step 1: Seed Question Collection \& Rule Abstraction}

We initialize the pipeline by collecting and transforming visual logic questions into structured \emph{seed rules} for scalable rule expansion. We first collect 4K visual logic questions together with their authored explanations from Chinese Civil Service Examination. Using LLM, we translate all explanations into English, rewrite them to eliminate answer leakage, and enrich them with clarifying details. Details for pre-processing steps can be found in our seeds dataset\footnote{\url{https://huggingface.co/datasets/VisualSphinx/VisualSphinx-Seeds}}. 
We then manually deduplicate the dataset and perform a consistency check by feeding each image–explanation pair to an LLM to verify correct answer retrieval.
This process leads to a filtered set of \textbf{2.4K} high-quality \emph{seed questions}. 
Each of these remaining questions is then abstracted into a structured \emph{seed rule} using an LLM, comprising five concise bullet points capturing the core visual pattern. 
Finally, the 2.4K seed rules are categorized into 8 classes with an LLM based on their visual patterns and required reasoning skills.
These classes specify the label space of subsequent evolutionary sampling. \textcolor{black}{Detailed descriptions of 8 classes are provided in Appendix~\ref{appendix: classification}.} \textcolor{black}{The detailed prompts for rewriting, verification, rule abstraction, and categorization can be found in Appendix~\ref{appendix: Prompt For Step 1}.}

The Chinese Civil Service Examination materials serve only as seed material for rule abstraction. The released \dataname{} training corpus is not a direct redistribution of the original exam images or authored explanations: after translation, rewriting, rule abstraction, and rule expansion, all training puzzles are rendered programmatically from generated Python code. The resulting dataset is released under the CC-BY-NC 4.0 license.

\begin{figure}[t]
  \centering
  \begin{minipage}[b]{0.5\textwidth}
    \centering
    \includegraphics[width=\textwidth]{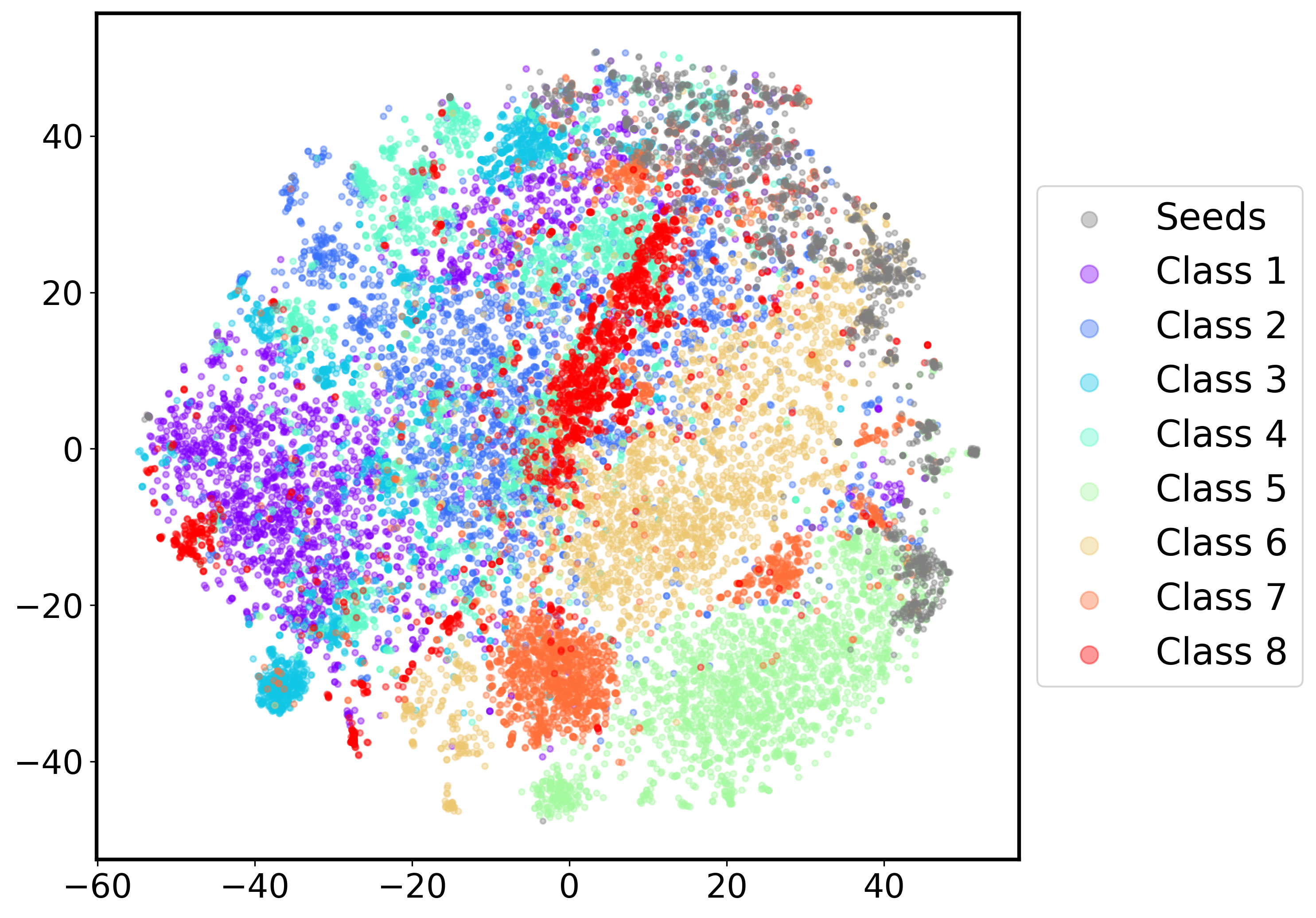}
    \caption{figure}{Comparison of t-SNE visualization between synthetic rules and seed rules with 2,000 sampled rules per class. Details about 8 classes are provided in Appendix~\ref{appendix: classification}.}
    \label{fig:tsne}
  \end{minipage}
  \hfill
  \begin{minipage}[b]{0.49\textwidth}
    \centering
    \includegraphics[width=\textwidth]{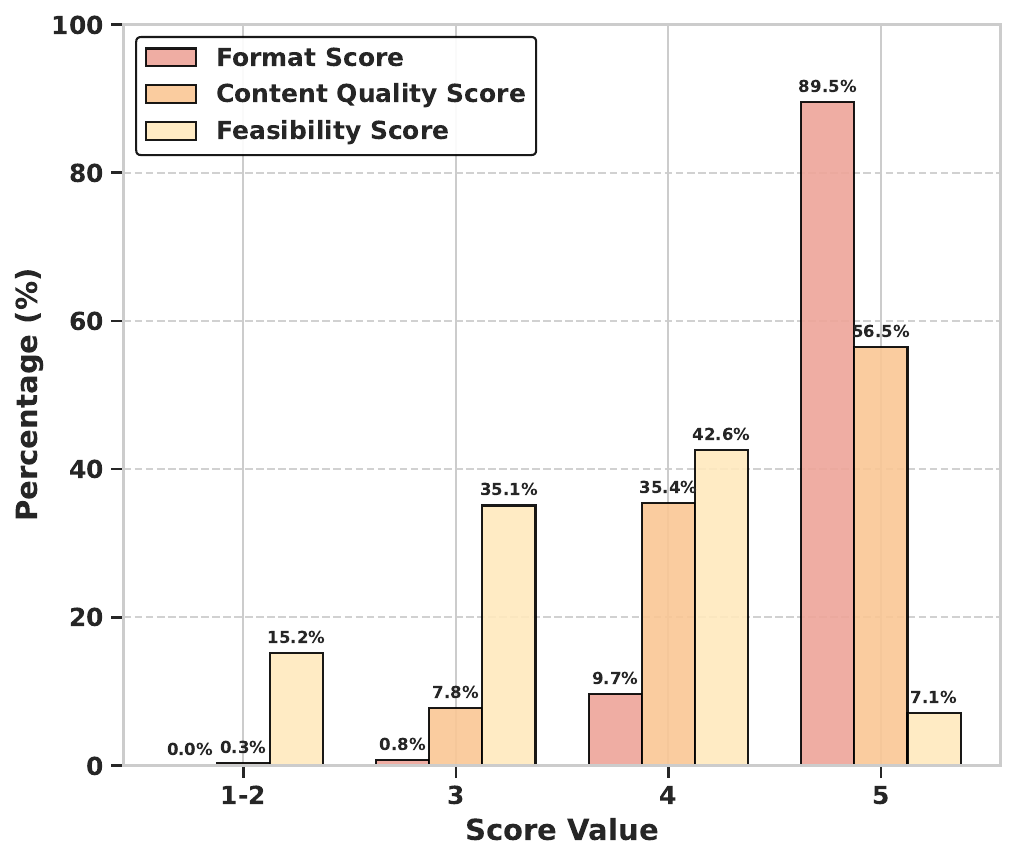}
    \caption{figure}{The statistics of synthetic rules format, content quality and feasibility score.}
    \label{fig:score}
  \end{minipage}
\end{figure}

\subsection*{Step 2: Rule Expansion via Genetic Algorithm}

\textbf{Rule Expansion.} To scale up the 2.4K \emph{seed rules}, we introduce a \emph{rule‑level genetic algorithm}. Each class of \emph{seed rules} forms a subpopulation, evolving independently on separate \emph{islands} through genetic operations: Mutation rewrites, adds, or deletes individual bullets, while crossover interleaves bullets from two parent rules. Every three generations, 10\% of the rules migrate across islands to maintain diversity. After ten generations, we obtain \textbf{60K} candidate rules. The detailed prompts for crossover and mutation can be found in Appendix~\ref{appendix: Prompt For Rule Expansion}.

To evaluate the diversity of generated rules in this step, we perform t-SNE~\citep{van2008visualizing} on the 60K candidate rules before final filtering to 40K, as this provides a comprehensive view of the rule space prior to pruning. We encode all rules using the all-mpnet-base-v2 sentence embedding model\footnote{\url{https://huggingface.co/sentence-transformers/all-mpnet-base-v2}} 
and apply t-SNE for dimensionality reduction, as shown in Figure~\ref{fig:tsne}. We observe that \dataname{}'s synthetic rules (shown in color) exhibit a broader distribution than seed rules (in gray), which cluster primarily in the upper right region, demonstrating a broad and well-separated distribution across the rule space. This analysis confirms that our genetic algorithm effectively diversifies the rule set, enhancing the variety of logical patterns available for puzzle generation.

\textbf{Rule Filtering.} To ensure these rules are meaningful and suitable for image synthesis in the subsequent step, we implement a rigorous evaluation and filtering process. First, we use the all-mpnet-base-v2 embedding model to project all rules into an embedding space and compute nearest-neighbor distances using FAISS~\citep{douze2024faiss}. Rules exceeding a predefined similarity threshold with existing entries are removed to eliminate redundancy and promote diversity. Deduplication details are in Appendix~\ref{appendix: deduplication}. Second, we use an LLM to rank the remaining rules based on three criteria: \textbf{format} (adherence to the structured five-bullet-point template), \textbf{content quality} (clarity and logical coherence of the rule), and \textbf{feasibility for code generation} (suitability for generating Python scripts to render images). Each criterion is scored from 1 to 5, and we retain only the top 40K rules that achieve a total score above 12 and a feasibility score of at least 3. \textcolor{black}{The score distribution can be found in Figure \ref{fig:score}, and the prompt we used can be found in Appendix~\ref{appendix: Prompt For Scoring}}. This filtering ensures that the selected rules are not only diverse but also interpretable and practically viable for generating visually distinct and logically consistent puzzles, thereby facilitating effective image synthesis in the following step.

\subsection*{Step 3: Program Based Rule‑to‑Image Synthesis}

\textbf{Rule‑to‑Image Synthesis.} For each of the 40K retained rules from Step 2, an LLM is prompted to generate two Python scripts in a single turn: \texttt{correct\_script.py} renders five sequential images that adhere to the rule, while \texttt{incorrect\_script.py} produces three plausible but rule-violating distractors. \textcolor{black}{The prompt template can be found in Appendix~\ref{appendix: Prompt For Step 3}.} Each rule yields one \textbf{image group}, containing a total of eight images (five correct and three incorrect). Within each group, four correct images serve as the question stem, the fifth correct image is the correct answer option, and the three incorrect images serve as distractors. These groups are used for puzzle assembly in Step 4.

\textbf{Image Style Diversification.} To enhance visual diversity and expand the dataset scale, \dataname{} employs three distinct rendering styles, each leveraging unique combinations of Python libraries, including \emph{Matplotlib}, \emph{PIL}, and \emph{NumPy}, to create varied graphical compositions. For each of the 40K rules from Step 2, we generate images in all three styles, tripling the dataset to approximately \textbf{120K} image groups. This approach encourages models to focus on abstract relational patterns rather than overfitting to specific visual features. Prompt templates of different rendering styles can be found in Appendix~\ref{appendix: Prompt For Step 3}.

\textbf{De-duplication.} To ensure the quality and uniqueness of the \textbf{120K} image groups, particularly to avoid identical images appearing among the question, correct option, or distractors, we implement a rigorous deduplication and validation pipeline. We use Perceptual Hashing (pHash)~\citep{hadmi12012perceptual} to compute hash values for each image within a group, measuring Hamming distances between all pairs. Images with distances below 10 are removed as duplicates. To identify low-quality images (e.g., those whose figures are too small), we compute the Structural Similarity Index (SSIM)~\citep{wang2004image} against a white reference image, flagging images with scores below 0.1 as blank, and calculate grayscale gradient energy, discarding images below a calibrated threshold. Groups with any image failing these criteria are removed, while those with high Hamming distances are retained if otherwise valid, as they may represent meaningful puzzles. This pipeline reduces the dataset to \textbf{110K} groups.

\subsection*{Step 4: Puzzle Assembly}

\textbf{Default Assembly.} From each verified group, our method constructs a four-option puzzle: the first four correct images, along with a question mark, constitute the question stem; the fifth correct image serves as the correct answer, while the three incorrect images are used as distractor options. All four options are randomly shuffled. The objective of the puzzle is to select the image that best completes the pattern indicated by the stem.

The designated answer is fixed before option shuffling and tracked through each permutation. We use \emph{answer-verifiable} to describe this deterministic traceability through the positive and negative generation programs, supported by automatic and human quality assessments. Program execution alone does not establish semantic correctness or answer uniqueness; these properties are assessed through the quality checks described in Section~\ref{sec: dataset statistics and analysis}.

\textbf{Alternative Assembly I: Answer Shuffling.} To increase diversity and mitigate position bias, we introduce a shuffled answer variant. For each image group, we construct four puzzles with identical question stems but different answer orderings, such that the correct image appears in options A, B, C, or D, respectively. This approach enriches training diversity by discouraging models from learning positional patterns. Only one variant is sampled per training instance to prevent memorization.

\textbf{Alternative Assembly II: Expanded Distractor Set.} To enhance difficulty and promote robust visual reasoning, we introduce a \textit{10-option puzzle variant}. Instead of using only the three incorrect distractors generated within the image group, we augment the option set with six additional distractors sampled from two other image groups associated with genetically related rules (e.g., parent or ancestor rules from Step 2’s rule-level genetic algorithm, rendered in the same code style). This creates puzzles with one correct option and nine plausible but rule-violating distractors, forming a challenging selection task.

Combining all puzzle construction strategies—including the default assembly, answer shuffling, and distractor expansion—the pipeline generates the \dataname{} dataset, comprising over \textbf{660K vision-logic puzzles} and providing a rich and diverse training corpus for multimodal model RL. The license of \dataname{} is CC-BY-NC 4.0.

%% file: sections/04-analysis.tex
\section{Dataset Analysis}

\begin{wrapfigure}{r}{0.4\textwidth}
    \centering
    \vspace{-2em}
    \includegraphics[width=0.4\textwidth]{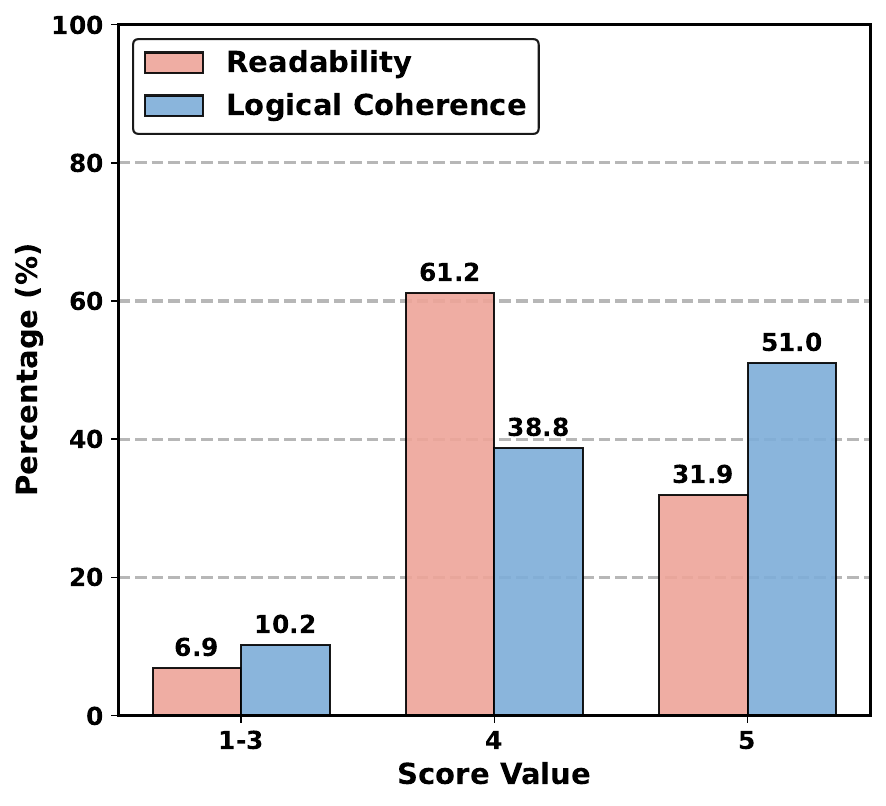}
    \caption{This figure demonstrates the statistics of readability and logical coherence.}
    \label{fig:readabilty and logical}
    \vspace{-3em}
\end{wrapfigure}

\label{sec: dataset statistics and analysis}
In this section, we conduct a comprehensive analysis to demonstrate the effectiveness of \dataname{} pipeline in generating high-quality, readable and logically coherent visual puzzles, as well as the low cost of generating the dataset.

\subsection{Dataset Attribute Analysis}

We analyze the quality and difficulty of the 660K synthetic puzzles in \dataname{} using three evaluation attributes: \textbf{readability}, \textbf{logical coherence}, and \textbf{pass rate}. 

\textbf{Attribute: Readability.} This metric evaluates the visual clarity and layout of puzzles generated in Step 4. It assesses whether the spatial arrangement, image scaling, or aspect ratios of the puzzle grid hinder understanding. An LLM is prompted to rate each puzzle’s readability on a scale of 1 to 5, with the score distribution shown in Figure~\ref{fig:readabilty and logical} (red). Results indicate that 93.1\% of the 660K puzzles achieve a score of 4 or higher, reflecting high visual clarity and robust layout fidelity across the large-scale dataset. \textcolor{black}{The prompts for annotation are in Appendix~\ref{appendix: Prompt for Annotation}.}

\textbf{Attribute: Logical Coherence.} A separate VLM, GPT-4.1-mini, receives the rendered puzzle, its intended rule, and its designated answer, and assigns a score from 1 to 5 based on rule--answer consistency, ambiguity, and answer uniqueness. Providing the rule and answer allows the evaluator to check the rendered puzzle against its generation specification. The score distribution, presented in Figure~\ref{fig:readabilty and logical} (blue), shows that 89.8\% of puzzles achieve a score of 4 or higher. This is an assessment conditioned on the intended rule and answer, rather than an independent blind-solving accuracy or a formal correctness guarantee. \textcolor{black}{The scoring rubric and prompts are in Appendix~\ref{appendix: Prompt for Annotation}.}

\begin{wrapfigure}{r}{0.4\textwidth}
    \centering
    \vspace{0em}
    \includegraphics[width=0.4\textwidth]{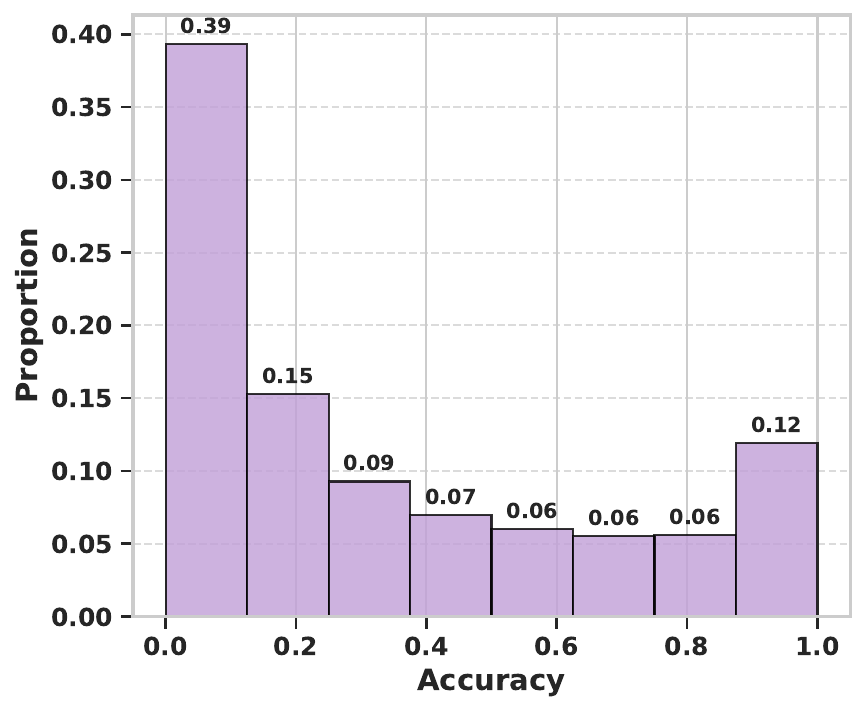}
    \caption{This figure demonstrates the statistics of pass rates.}
    \label{fig:passrate}
    \vspace{-1em}
\end{wrapfigure}

\textbf{Attribute: Pass Rate.} We measure the pass rate using a VLM.
We note that current open-source VLMs not trained on these tasks struggle to achieve useful pass rates on visual logic puzzles and perform close to random guessing. To evaluate the solvability of the generated puzzles, we apply GRPO \citep{shao2024deepseekmath} to train an annotation model based on \textsc{Qwen2.5-VL-7B}, using our filtered seed questions. We then evaluate this model on a full set of 110,000 filtered image groups (generated using default assembly). Detailed training hyperparameters are provided in Appendix~\ref{appendix: experimental setups for RL training}. The resulting pass rates are shown in Figure~\ref{fig:passrate}. Notably, 39\% of puzzles, or roughly 43K default-assembly puzzles, are never correctly solved (accuracy of 0), indicating that a significant portion of the dataset poses novel and non-trivial challenges for the model. This highlights the dataset’s potential for testing generalization and reasoning capabilities in vision-based logical reasoning tasks. The remaining pass rates are distributed relatively evenly across difficulty bins, reflecting a diverse range of challenge levels and further validating the utility of the \dataname{} dataset for RL training.

\textbf{More Analysis.}
We defer detailed human-based quality analysis to Appendix \ref{appendix: human analysis}. Manual evaluation by human annotators confirms \dataname{}'s high quality, complementing automated VLM-based assessments.

\subsection{Cost \& Model Selection Analysis}
Our pipeline generates the \dataname{} of over 660K high-quality visual logic puzzles at a total cost of less than \$1000, averaging approximately \$0.0015 per puzzle. We select models for each step based on their reasoning capabilities, modality requirements, and cost efficiency, ensuring optimal performance while maintaining low costs. More details for model selection can be found in Appendix~\ref{appendix: model selection}.

\textbf{Step 1: Seed Question Processing.} We employ Claude-3.5-Haiku and Claude-3.7-Sonnet \citep{claude3} for multimodal tasks including translating and rewriting explanations, verifying image-text answers, abstracting rules, and classifying 2.4K seed questions. Each question is processed with four prompts at a cost of \$8 per 1K questions, totaling approximately \$77 for 2.4K questions.

\textbf{Step 2: Rule Expansion.} We use DeepSeek-V3 \citep{liu2024deepseek}, a text-based model, for rule expansion via a rule-level genetic algorithm, scaling 2.4K seed rules to 60K synthetic rules. At \$2 per 1K rules, the cost totals approximately \$120.

\textbf{Step 3: Image Synthesis.} We utilize Grok-3-mini \citep{xai2025grok3} with high reasoning effort to generate two Python scripts per rule—one rendering five correct images, another producing three distractors—ensuring consistent script generation for 120K image groups. At \$6 per 1K groups, the cost is approximately \$720.

\textbf{Step 4: Puzzle Assembly and Evaluation.} For quality assurance, we evaluate 110K puzzles assembled using the default assembly strategy. A fine-tuned Qwen2.5-VL-7B-Instruct model assesses pass rate, while GPT-4.1-mini, a lightweight vision-language model, evaluates visual readability and logical coherence at \$0.70 per 1K puzzles, totaling approximately \$77.

%% file: sections/05-exp.tex
\section{Performance Analysis of VisualSphinx on RL}
\label{sec: performance analysis}
In this section, we evaluate the quality of the \dataname{} datasets by using them to perform RL on vision models, assessing their effectiveness via benchmark performance.

\subsection{Experimental Setups}
\label{section: experiments setups}

\textbf{\dataname{} Setup.} Given the large size of the \dataname{} dataset, we sample 10,000 challenging yet feasible examples, selecting those with pass rates between 0.375 and 0.875, and a combined readability and logical coherence score (as defined in Section~\ref{sec: dataset statistics and analysis}) of 8 or higher. To ensure sufficient diversity and challenge, we include 80\% of the questions with 4 options and the remaining 20\% with 10 options. This setup validates whether a high-quality subset of \dataname{} provides useful RL signal; we leave a full study of performance as a function of dataset scale to future work.

\textbf{RL Setup.} We train \textsc{Qwen2.5-VL-3B} and \textsc{Qwen2.5-VL-7B} \citep{qwen2.5-VL} using GRPO~\citep{shao2024deepseekmath}, with a rollout batch size and PPO mini-batch size of 64 and 8 responses per sample. All reported benchmark results for these models use their step-256 checkpoints. The 7B run was continued to step 320 to observe the optimization trajectory shown in Figure~\ref{fig:accuracy_subplots}. A binary reward of 1 is assigned when the model's final answer matches the designated answer, and 0 otherwise. This reward is computed by answer matching without an additional judge model or formatting reward. Training is conducted on a cluster of 8 H100 GPUs with 2TB RAM. The training hyperparameters are provided in Appendix~\ref{appendix: experimental setups for RL training}.

\textbf{SFT Setup.} To compare our RL setup with supervised fine-tuning (SFT) in Section~\ref{section: ablation}, we perform full-parameter SFT of \textsc{Qwen2.5-VL-7B-Instruct} on two datasets constructed from \dataname{}, each containing 10{,}000 puzzles. In \emph{SFT-Random}, we randomly sample 10{,}000 puzzles and pair each with one response generated by \textsc{Qwen2.5-VL-7B-Instruct}. In \emph{SFT-Reject}, we sample up to 5 responses per puzzle from \textsc{Qwen2.5-VL-32B-Instruct} and retain the first response whose final answer matches the designated answer \citep{touvron2023llama2openfoundation}; the final dataset contains 10{,}000 retained examples. Training is conducted on 8 H100 GPUs. The training hyperparameters are detailed in Appendix~\ref{appendix: experimental setups for sft}.

\textbf{Benchmarks.} To validate the effectiveness of improving the performance of vision-based logical reasoning, we construct \dataname{}-\textsc{Bench}, a test set of 930 high-quality visual logic puzzles from the \dataname{} dataset. Each puzzle is manually checked for correctness and high quality, comprising both 4-option and 10-option questions. The 930 puzzles are categorized into three difficulty levels based on their pass rates: easy (pass rates between 0.5 and 0.75), medium (pass rates between 0.25 and 0.5), and hard (pass rate of 0). We emphasize that this test set is fully synthetic and has no instance-level overlap with existing benchmarks, training datasets, or our sampled 10K RL training data. \dataname{}-\textsc{Bench} is therefore an in-distribution held-out benchmark for the proposed \dataname{} visual-logic distribution, while seed-rule-disjoint evaluation separately tests extrapolation to rule families whose ancestral seed rules are not used for training.

\input{tables/benchmark}

\input{tables/eval}

In addition to \dataname{}-\textsc{Bench}, we evaluate RL-trained models on a suite of benchmarks spanning multiple domains: \textbf{external visual-logic} (VisuLogic~\citep{xu2025visulogic} and RAVEN~\citep{zhang2019raven}), \textbf{vision-math} (MathVerse-vision~\citep{zhang2024mathverse}, MathVista-mini~\citep{lu2023mathvista}, We-Math-mini~\citep{qiao2407we}), \textbf{vision-general} (MMMU-pro standard~\citep{yue2024mmmu}), \textbf{vision-coding} \citep{wang2025code} (Algorithm, HumanEval-V, MATH), and \textbf{pure text-math} (AIME25, AMC, Olympiad \citep{he2024olympiadbench}, MATH-500 \citep{hendrycks2021measuring}). This evaluation protocol measures both held-out performance on the proposed \dataname{} distribution and benchmark-dependent transfer to external reasoning tasks. Each reported training configuration is run once, and all benchmark evaluations use temperature-0 greedy decoding. Greedy decoding removes response-sampling randomness but does not quantify variation from training, data sampling, or checkpoint selection. We report observed accuracy differences without claiming statistical significance across training seeds. Detailed descriptions of these benchmarks and answer extraction rules are provided in Appendix~\ref{appendix: evaluation details} and Appendix~\ref{appendix: benchmark details}.

\subsection{Experimental Results}
\label{section: experiments results}

\textbf{\dataname{} is Effective in Increasing the Logical Reasoning Ability of VLMs.} Table~\ref{tab:benchmark} and Figure~\ref{fig:accuracy_subplots} compare the performance of models before and after RL training with \dataname{} on \dataname{}-\textsc{Bench} across difficulty levels. The RL-trained models substantially improve their accuracy across all difficulty levels. The 7B model reaches GPT-4.1-comparable overall performance, with stronger performance on Easy examples but lower performance on Medium and Hard examples. These results support the utility of \dataname{} for RL training on visual logic puzzles.

\textbf{\dataname{} Generalizes to Seed-Rule-Disjoint Examples.} We trace the evolutionary provenance of the benchmark rules and identify 71 examples whose ancestral seed-rule sets have no overlap with the ancestral seed rules of the 10K RL training set. Appendix~\ref{appendix: evaluation details} describes this criterion. On this subset, the RL-trained 3B and 7B models improve over their Instruct baselines by 25.36 and 28.17 percentage points, respectively; detailed results are provided in Appendix Table~\ref{tab:seed_disjoint}. These gains are comparable to those on the full benchmark and provide evidence of transfer to rule lineages with held-out ancestral seeds.

\textbf{\dataname{} Transfers to External Visual-Logic Benchmarks.} Table~\ref{tab:model_comparison} includes two external visual-logic benchmarks, VisuLogic and RAVEN, which differ from \dataname{} in puzzle format and visual structure. On VisuLogic, the gains are modest but positive: \(+1.50\%\) for 3B and \(+2.10\%\) for 7B. On RAVEN, the average improvements are \(+5.14\%\) for 3B and \(+5.90\%\) for 7B, with layout-level details provided in Appendix Table~\ref{tab:raven_details}. These results provide external visual-logic evidence that \dataname{} training can improve visual pattern reasoning beyond the exact generation pipeline.

\textbf{\dataname{} Shows Benchmark-Dependent Transfer Beyond Visual Logic.} Table~\ref{tab:model_comparison} presents a comparison of model performance before and after RL training on \dataname{}, evaluated across both 7B and 3B model scales. For the 7B model, RL training improves vision-math benchmarks by +6.98\% (MathVerse), +5.30\% (MathVista), and +1.09\% (We-Math); the in-domain \dataname{}-\textsc{Bench} improves by +26.64\%; vision-general reasoning (MMMU-pro) increases by +2.55\%; and vision-coding tasks improve by +1.34\% (Algorithm), +3.05\% (HumanEval-V), and +3.20\% (MATH). Pure text-math benchmarks also show small positive gains of +3.33\% (AIME25), +2.50\% (AMC), +1.93\% (Olympiad), and +1.40\% (MATH-500). The 3B model shows strong gains on \dataname{}-\textsc{Bench} (+29.30\%) and several external tasks, but it does not improve uniformly; for example, MathVerse-vision decreases by 1.01\%. We therefore characterize the transfer as positive but benchmark-dependent, rather than uniformly broad across all domains and model sizes.

\subsection{Ablation Analysis}
\label{section: ablation}

\input{tables/ablation}

Table~\ref{tab:ablation_sft} compares GRPO with two practical SFT recipes on \dataname{}. Under the reported configurations, GRPO achieves higher accuracy than both SFT variants and the original Instruct baseline on all five benchmarks. On \dataname{}-\textsc{Bench}, GRPO reaches 55.94\%, while the stronger SFT variant reaches 49.52\%. Its gains over the stronger SFT variant on each benchmark are 4.57 percentage points (MathVerse-Vision), 7.50 (MathVista-mini), 10.40 (We-Math-mini), 6.42 (\dataname{}-\textsc{Bench}), and 6.65 (MMMU-Pro). SFT-Reject achieves higher in-domain accuracy than SFT-Random, while its results on the external math and general-vision benchmarks are mixed. All three training datasets contain 10{,}000 puzzles, but response supervision, token exposure, optimization updates, and compute budgets are not matched. These results demonstrate useful supervision for RL post-training under the evaluated setup; they do not isolate the causal effect of the training algorithm.

%% file: tables/benchmark.tex
\begin{table}[!t]
  \vspace{-2em}
  \centering
  \caption{Accuracy of vision models on \dataname{}-\textsc{Bench} across difficulty levels. \textsc{Qwen2.5-VL-3B-\dataname{}} and \textsc{Qwen2.5-VL-7B-\dataname{}} represent our RL-trained models fine-tuned on \dataname{}. The 7B model reaches GPT-4.1-comparable overall performance, performs strongest on Easy examples, and remains lower on Medium and Hard examples. Numbers with the best performance for each difficulty level are highlighted in \textbf{bold}, while those with the second-highest performance are \underline{underlined}.}
  \label{tab:benchmark}
  \resizebox{0.9\textwidth}{!}{
  \begin{tabular}{c l l l l  }
    \toprule
    \textbf{Model}          & \multicolumn{1}{l}{ \textbf{Overall (\%)}}  & \textbf{Easy (\%)} & \textbf{Medium (\%)} & \textbf{Hard (\%)} \\
    \midrule
    Grok-2-Vision& 32.73& 41.99& 30.97& 25.24\\
    Gemini-1.5-Pro& 42.99& 50.32& 40.00& 38.66\\
    Gemini-2.0-Flash& 42.78& 50.00& 43.55& 34.82\\
    Gemini-2.5-Flash& 49.20& 58.01& 48.71& 40.89\\
    Gemini-2.5-Pro& 50.27& 60.26& 49.35& 41.21\\
    GPT-4o& 49.30& 55.45& 49.68& \underline{42.81}\\
    GPT-4.1& \underline{55.72}& 66.67& \textbf{54.19}& \textbf{46.33}\\
    o4-mini& 41.93& 44.87& 47.74& 33.22\\
    Claude-3.5-Haiku& 35.72& 45.51& 34.52& 27.16\\
    Claude-3.7-Sonnet& 48.88& 55.77& 49.03& 41.85\\
    Claude-4-Sonnet& 46.31& 52.24& 44.19& 42.49\\
    Claude-4-Opus& 50.48& 54.17& 50.98& \textbf{46.33}\\
    \midrule
    \textsc{Qwen2.5-VL-3B}& 23.53& 41.71& 21.41& 16.52\\
    \textsc{Qwen2.5-VL-3B-\dataname{}}& 52.83 \tiny{\textcolor{red}{+29.3}}& \underline{72.33} \tiny{\textcolor{red}{+30.62}}& 49.35 \tiny{\textcolor{red}{+27.94}}& 34.82 \tiny{\textcolor{red}{+18.3}}\\
    \textsc{Qwen2.5-VL-7B}& 29.30& 42.31& 23.87& 21.73\\
    \textsc{Qwen2.5-VL-7B-\dataname{}} & \textbf{55.94} \tiny{\textcolor{red}{+26.64}}& \textbf{76.92} \tiny{\textcolor{red}{+34.61}}& \underline{52.26} \tiny{\textcolor{red}{+28.39}} & 38.66 \tiny{\textcolor{red}{+16.93}}\\ 
    \bottomrule
  \end{tabular}
  }
  \par\vspace{1em}
  \begin{minipage}{\textwidth}
    \centering
    \captionsetup{type=figure}
    \includegraphics[width=0.9\textwidth]{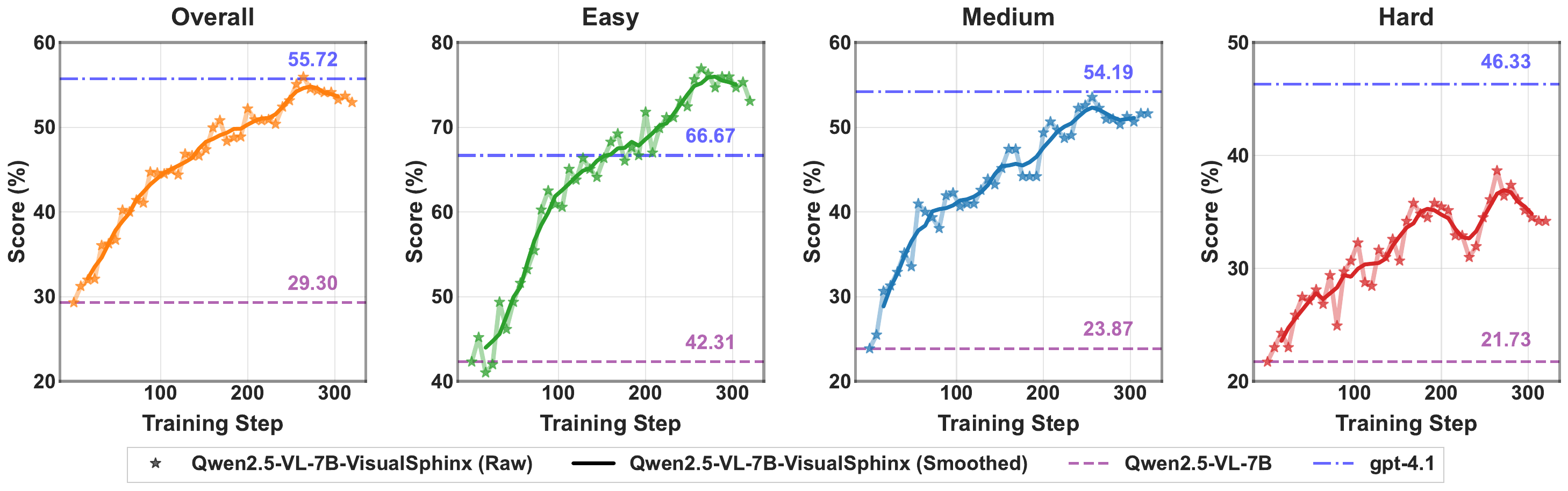}
    \caption{Performance of \textsc{Qwen2.5-VL-7B} trained on \dataname{}, evaluated on \dataname{}-\textsc{Bench} across difficulty levels. The curves include continued training through step 320 to illustrate the optimization trajectory; all tabulated benchmark results for the RL-trained models use the step-256 checkpoints.}

    \vspace{-2em}
    \label{fig:accuracy_subplots}
  \end{minipage}
\end{table}

%% file: tables/eval.tex
\begin{table}[t]
  \vspace{-2em}
  \centering
  \caption{Performance on different benchmarks (accuracy, \%). RAVEN reports the average over seven layouts; layout-level results are provided in Appendix Table~\ref{tab:raven_details}.}
  \vspace{1em}
  \label{tab:model_comparison}
  \small
  \setlength{\tabcolsep}{5pt}
  \resizebox{\textwidth}{!}{
  \begin{tabular}{@{}l *{4}{c}@{}}
    \toprule
    \textbf{Benchmark} & \textbf{Qwen2.5-VL-7B} & \textbf{Qwen2.5-VL-7B} & \textbf{Qwen2.5-VL-3B} & \textbf{Qwen2.5-VL-3B} \\
                       & \textbf{Instruct} & \textbf{\dataname{}} & \textbf{Instruct} & \textbf{\dataname{}} \\
    \midrule
    \multicolumn{5}{@{}l}{\textit{\textbf{Vision -- Math}}} \\
    \addlinespace[0.2em]
    Mathverse vision & 27.79 & \textbf{34.77} & 23.35 & 22.34 \\
    Mathvista mini & 60.00 & \textbf{65.30} & 33.60 & 47.60 \\
    We-Math mini & 61.21 & \textbf{62.30} & 54.20 & 59.20 \\
    \addlinespace[0.5em]
    \midrule
    
    \multicolumn{5}{@{}l}{\textit{\textbf{Vision -- Puzzle/Logic}}} \\
    \addlinespace[0.2em]
    \dataname{}-\textsc{Bench} & 29.30 & \textbf{55.94} & 23.53 & 52.83 \\
    VisuLogic & 24.40 & \textbf{26.50} & 24.90 & 26.40 \\
    RAVEN (avg.) & 19.99 & \textbf{25.89} & 13.16 & 18.30 \\
    \addlinespace[0.5em]
    \midrule

    \multicolumn{5}{@{}l}{\textit{\textbf{Vision -- General}}} \\
    \addlinespace[0.2em]
    MMMU-pro standard & 50.46 & \textbf{53.01} & 34.28 & 43.35 \\
    \addlinespace[0.5em]
    \midrule

    \multicolumn{5}{@{}l}{\textit{\textbf{Vision -- Coding}}} \\
    \addlinespace[0.2em]
    Algorithm & 51.68 & \textbf{53.02} & 33.56 & 36.91 \\
    HumanEval-V & 78.66 & \textbf{81.71} & 62.80 & 64.63 \\
    MATH & 69.60 & \textbf{72.80} & 56.80 & 60.80 \\
    \addlinespace[0.5em]
    \midrule

    \multicolumn{5}{@{}l}{\textit{\textbf{Pure Text -- Math}}} \\
    \addlinespace[0.2em]
    AIME25 & 0.00 & \textbf{3.33} & 0.00 & \textbf{3.33} \\
    AMC & 32.50 & \textbf{35.00} & 20.00 & 27.50 \\
    Olympiad & 31.11 & \textbf{33.04} & 23.41 & 25.48 \\
    MATH-500 & 64.40 & \textbf{65.80} & 54.60 & 56.20 \\
    \bottomrule
  \end{tabular}
  }
\end{table}

%% file: tables/ablation.tex
\begin{table}[t]
  \vspace{-2em}
  \centering
  \caption{Ablation on \dataname{}: SFT vs.\ RL across five benchmarks (accuracy, \%).}
  \vspace{0.5em}
  \label{tab:ablation_sft}
  \begin{tabular}{@{}l *{4}{c}@{}}
    \toprule
    \textbf{Benchmark} & \textbf{Qwen2.5-VL-7B} & \textbf{SFT} & \textbf{SFT} & \textbf{\textsc{Qwen2.5-VL-7B}} \\
                       & \textbf{Instruct} & \textbf{Random} & \textbf{Reject} & \textbf{\dataname{}} \\
    \midrule
    \multicolumn{5}{@{}l}{\textit{\textbf{Vision -- Math}}} \\
    \addlinespace[0.3em]
    MathVerse vision   & 27.79 & 30.20 & 23.35 & \textbf{34.77} \\
    Mathvista mini     & 60.00 & 56.70 & 57.80 & \textbf{65.30} \\
    We-Math mini       & 61.21 & 51.90 & 44.48 & \textbf{62.30} \\
    \addlinespace[0.5em]
    \midrule
    
    \multicolumn{5}{@{}l}{\textit{\textbf{Vision -- Puzzle}}} \\
    \addlinespace[0.3em]
    \dataname{}-\textsc{Bench} & 29.30 & 39.89 & 49.52 & \textbf{55.94} \\
    \addlinespace[0.5em]
    \midrule
    
    \multicolumn{5}{@{}l}{\textit{\textbf{Vision -- General}}} \\
    \addlinespace[0.3em]
    MMMU-pro standard  & 50.46 & 46.36 & 46.01 & \textbf{53.01} \\
    \bottomrule
  \end{tabular}
  \vspace{-1em}
\end{table}

%% file: sections/07-conclusion.tex
\section{Conclusion and Future Work}
\label{sec: conclusion}

In this work, we introduced \dataname{}, a pioneering large-scale synthetic dataset comprising over 660K visual logic puzzles designed to enhance the logical reasoning capabilities of vision-language models (VLMs) through reinforcement learning (RL). Our four-stage pipeline leverages rule abstraction, rule-level genetic algorithms, program-based image synthesis, and strategic puzzle assembly to generate diverse and interpretable visual reasoning tasks at a cost of less than \$1000. Our empirical results show that fine-tuning \textsc{Qwen2.5-VL-7B} with a standard GRPO recipe substantially improves accuracy on \dataname{}-\textsc{Bench}, with smaller gains on external visual-logic benchmarks and positive but benchmark-dependent transfer to other visual reasoning tasks.

\dataname{} addresses a critical gap in the availability of large-scale, structured datasets for multimodal reasoning. Future work could study performance scaling as a function of dataset size, extend \dataname{} to more complex reasoning paradigms such as temporal or interactive tasks, and investigate the theoretical underpinnings of its effectiveness.

%% file: sections/appendix.tex
\clearpage
\input{sections/appendix01}

\input{sections/appendix02}

\input{sections/appendix03}

\input{sections/appendix04}

\input{sections/appendix05}

\input{sections/appendix06}

%% file: sections/appendix01.tex
\section{More Examples and \dataname{} Statistics}
\label{appendix: statistics}

\subsection{Examples}
In this section, we demonstrate more examples of \dataname{}.

\begin{figure}[H]
    \vspace{-0.5em}
    \centering
    \includegraphics[width=0.55\textwidth]{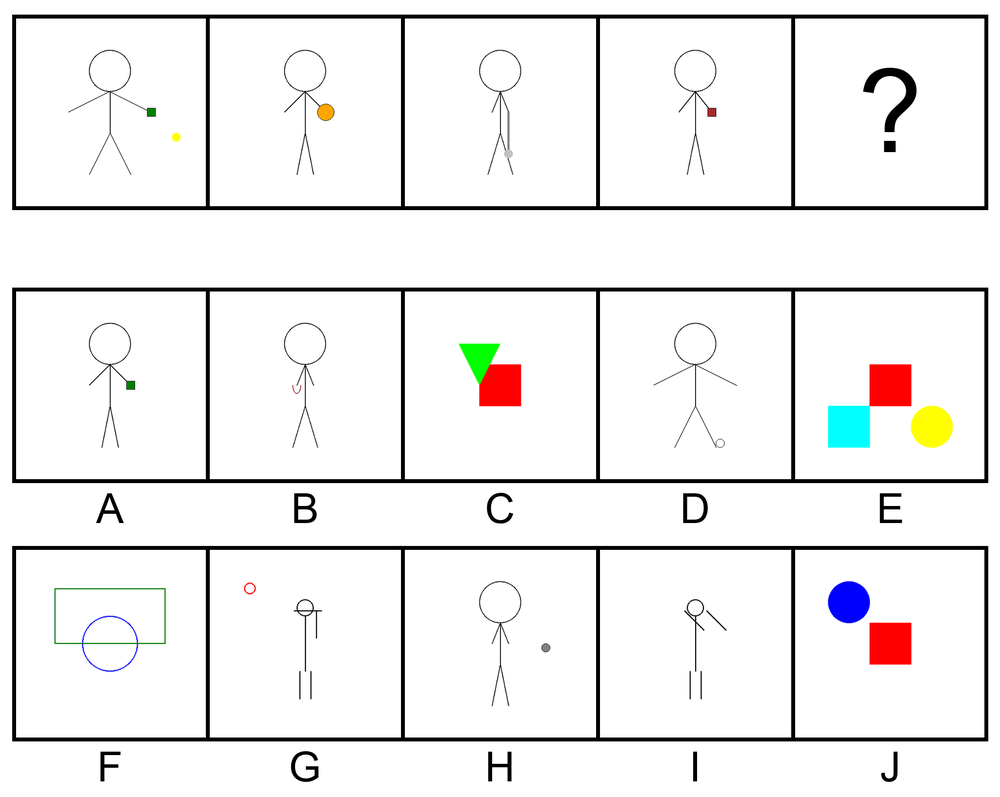}
    \caption{Example 1}
    \label{fig:example01}
\end{figure}

\textbf{Example 1 in Figure~\ref{fig:example01}} \\
\textit{Reasoning type:} Inductive (10 options) \\
\textit{Rule:} The stick figure is required to hold a solid object in its hand. \\
\textit{Answer:} A

\begin{figure}[H]
    \vspace{-3em}
    \centering
    \includegraphics[width=0.55\textwidth]{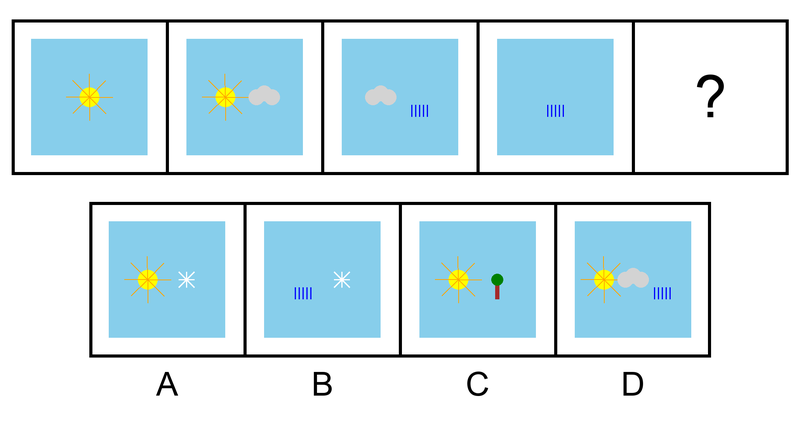}
    \caption{Example 2}
    \label{fig:example02}
\end{figure}

\textbf{Example 2 in Figure~\ref{fig:example02}} \\
\textit{Reasoning type:} Deductive (4 options) \\
\textit{Rule:} Each weather icon in the images shifts progressively to the left across the sequence. \\
\textit{Answer:} B

\begin{figure}[H]
    \vspace{-2em}
    \centering
    \includegraphics[width=0.55\textwidth]{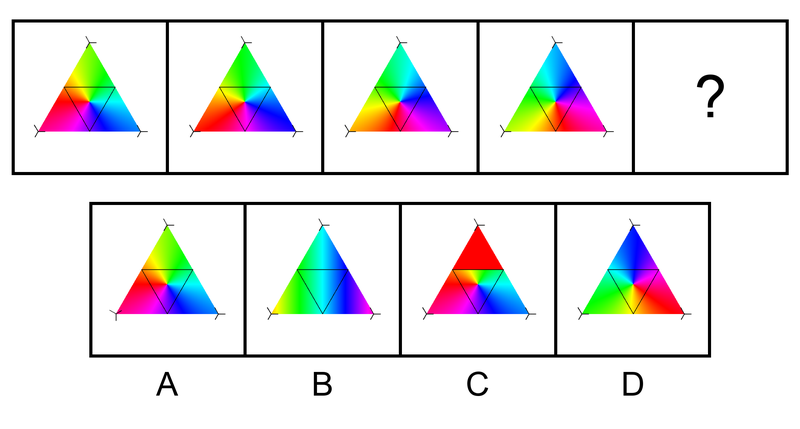}
    \caption{Example 3}
    \label{fig:example03}
\end{figure}

\textbf{Example 3 in Figure~\ref{fig:example03}} \\
\textit{Reasoning type:} Deductive (4 options) \\
\textit{Rule:} The green part in each image rotates counterclockwise in sequence. \\
\textit{Answer:} D

\begin{figure}[H]
    \vspace{-2em}
    \centering
    \includegraphics[width=0.55\textwidth]{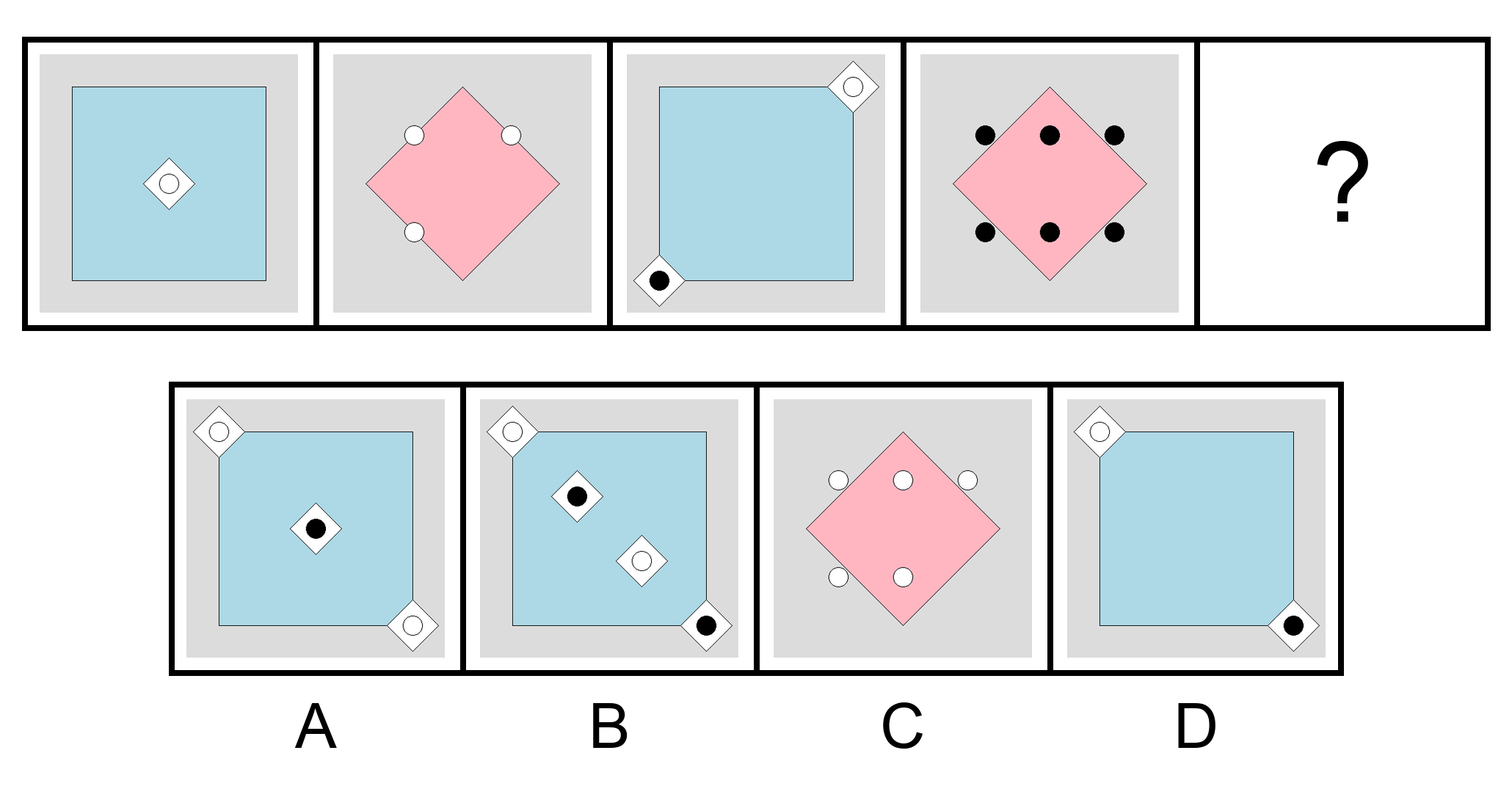}
    \caption{Example 4}
    \label{fig:example04}
\end{figure}

\textbf{Example 4 in Figure~\ref{fig:example04}} \\
\textit{Reasoning type:} Deductive (4 options) \\
\textit{Rule:} The number of balls doubles every other image in the sequence. \\
\textit{Answer:} B

\begin{figure}[H]
    \vspace{-2em}
    \centering
    \includegraphics[width=0.55\textwidth]{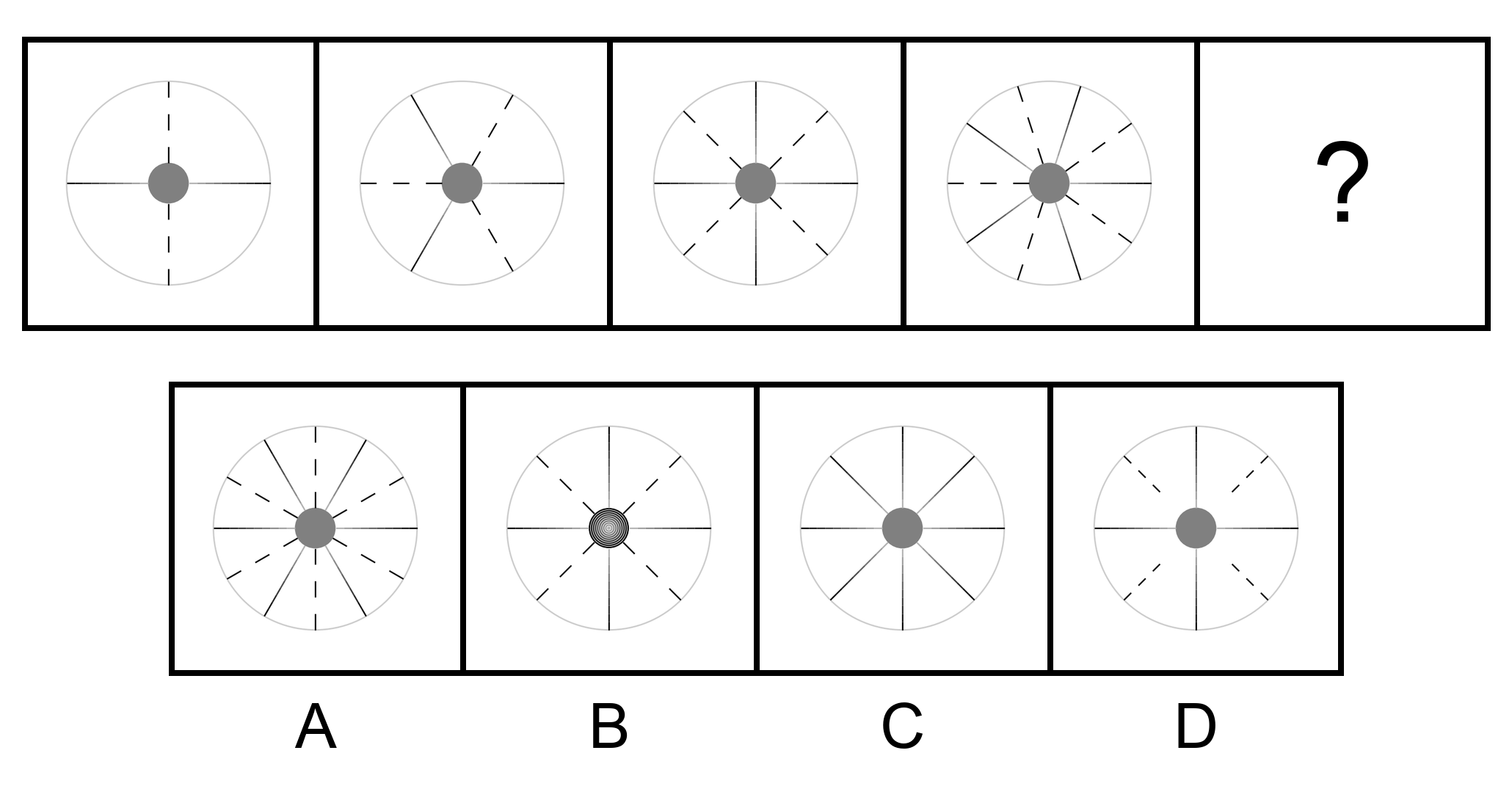}
    \caption{Example 5}
    \label{fig:example05}
\end{figure}

\textbf{Example 5 in Figure~\ref{fig:example05}} \\
\textit{Reasoning type:} Deductive (4 options) \\
\textit{Rule:} The number of solid and dashed lines doubles sequentially across the images. \\
\textit{Answer:} A

\subsection{Breakdown of Statistics of \dataname{}}

Table~\ref{tab:statistics} summarizes the data volume across all stages of the \dataname{} pipeline. Starting from 3,904 seed questions, we obtain 2,398 high-quality seed rules through filtering and verification. These are expanded to 60,339 candidate rules and further refined to 41,287 retained rules after de-duplication and LLM-based scoring.

Using three distinct rendering styles, we generate over 117K image groups, of which 110,423 remain after filtering out blank or degenerate outputs. From these remaining image groups, we construct puzzles employing three strategies: a default 4-option format, an answer-shuffling variant, and a 10-option hard-distractor variant.

\input{tables/statistics}

%% file: tables/statistics.tex
\begin{table}[H]
\centering
\caption{Detailed statistics at each stage of the \dataname{} pipeline.}
\vspace{1em}
\resizebox{\textwidth}{!}{
\begin{tabular}{l  r  r}
\toprule
\textbf{Stage} & \textbf{Output Count} & \textbf{Description} \\
\midrule
Seed Questions & 3,904 & Raw visual logic questions \\
Filtered Seed Rules & 2,398 & After rewriting and verification \\
Generated Rules & 60,339 & After rule-level evolution \\
Retained Rules & 41,287 & After deduplication and scoring \\
\midrule
Style 1 Image Groups & 39,307 & Successfully rendered \\
Style 2 Image Groups & 38,918 & Successfully rendered \\
Style 3 Image Groups & 39,360 & Successfully rendered \\
\textbf{Total Image Groups} & \textbf{117,585} & Across all three styles \\
\midrule
Filtered Image Groups & 110,423 & After blank removal \\
\midrule
Default Puzzle Assembly & 110,423 & 1 puzzle per image group \\
Answer Shuffling Variant & 441,692 & 4 puzzles (different option order) per group  \\
10-option Variant & 110,423 & 1 puzzle (10-option) per group \\
\midrule
\textbf{Total Puzzles} & \textbf{662,538} & Across all assembly strategies \\
\bottomrule
\end{tabular}
}
\label{tab:statistics}
\end{table}

%% file: sections/appendix02.tex
\section{Additional Analysis of Synthetic Rules}

\subsection{Categorization of Rules}
\label{appendix: classification}

\begin{figure}[!tb]
    \vspace{-1em}
    \centering
    \includegraphics[width=1\textwidth]{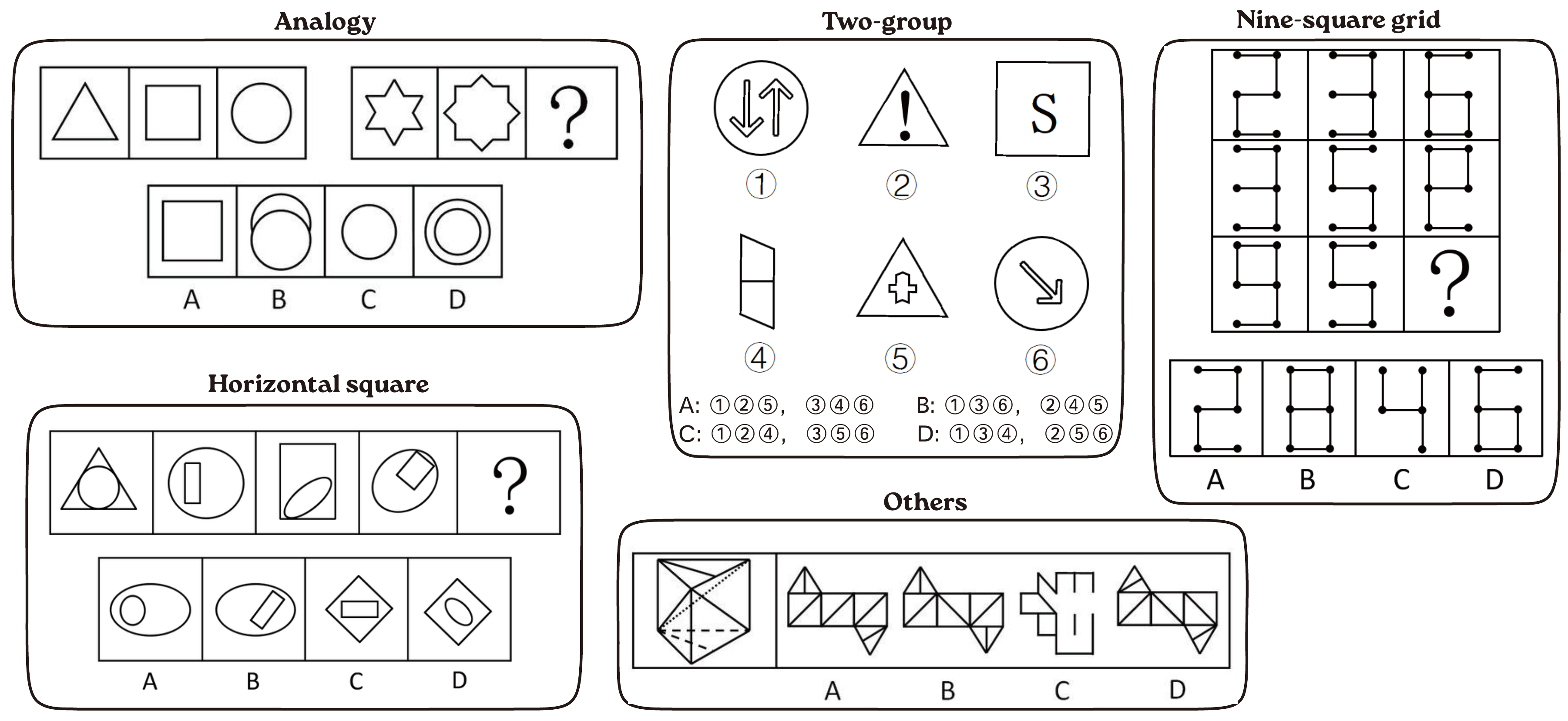}
    \caption{Five different visual patterns of seeds}
    \label{fig:rule-categories}
\end{figure}

We categorize seed rules along two axes: \textit{visual pattern} and \textit{reasoning style}. Figure~\ref{fig:rule-categories} illustrates representative examples from five visual patterns of seeds: \textit{Nine-square grid}, \textit{Horizontal square}, \textit{Analogy}, \textit{Two-groups}, and \textit{Others}. Reasoning styles are labeled as \textit{Deductive}, \textit{Inductive}, or \textit{Others}. We note that in principle, these two axes yield 15 possible visual pattern–reasoning style combinations. However, we collapse all combinations involving an \textit{Others} tag on either axis into a unified \textit{Others} category for simplicity. Additionally, rules classified as \textit{Two-group} pattern are all with \textit{Inductive} reasoning style. Taking these constraints into account, we arrive at 7 valid categories and a fallback \textit{Others} group, resulting in 8 total rule types used for island-based evolution.

Table~\ref{tab:rule-category-distribution} and the following \textbf{tags} summarize the distribution of seed rules across visual patterns and reasoning styles.

\input{tables/classes}

\textbf{Visual pattern tags:}
\begin{itemize}
    \item \textbf{Nine-square grid} — A $3 \times 3$ grid with one cell replaced by a question mark.
    \item \textbf{Horizontal square} — A row of 4 – 6 squares, one of which is missing or marked.
    \item \textbf{Analogy} — Two groups of three squares, one containing a question mark.
    \item \textbf{Two-group} — Images include numbers to be partitioned into two groups.
    \item \textbf{Others} — Any layout that does not fit the above definitions.
\end{itemize}

\textbf{Reasoning style tags:}
\begin{itemize}
    \item \textbf{Deductive} — Each image builds on the previous with a direct transformation or rule.
    \item \textbf{Inductive} — All images reflect a shared underlying rule without strict ordering.
    \item \textbf{Others} — Patterns that fall outside deductive or inductive reasoning.
\end{itemize}

This classification plays a critical role in our \emph{rule‑level genetic algorithm}, as combining points from incompatible visual patterns often results in semantic incoherence, even when LLMs are instructed to rewrite the rules according to context. For instance, a rule component specific to \textbf{Nine-square grid} may not generalize to \textbf{Two-group}. Such inconsistencies significantly reduce the success rate of downstream rule-to-image synthesis, often producing puzzles that fail to reflect the intended logic.

To address this challenge, we assign each rule category to a subpopulation that evolves independently on its own island, where genetic operations such as crossover and mutation are applied exclusively within groups sharing the same visual pattern and reasoning style. Every three generations, 10\% of the rules are randomly selected to migrate across islands, promoting diversity while preserving the local semantic structure. This island-based design effectively balances semantic coherence with exploratory variation, ensuring the generation of coherent and diverse logical puzzles for \dataname{}.

\subsection{Deduplication of Rules}
\label{appendix: deduplication}

\begin{wrapfigure}{r}{0.4\textwidth}
    \centering
    \vspace{-2em}
    \includegraphics[width=0.4\textwidth]{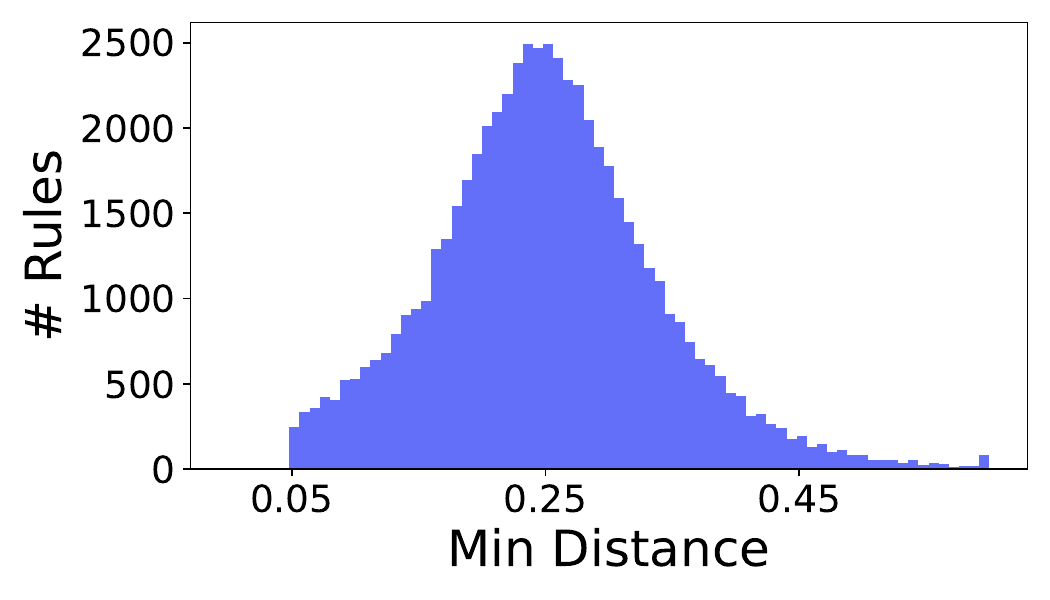}
    \caption{Rule distance.}
    \label{fig:min_distanc}
    \vspace{-2em}
\end{wrapfigure}

We quantify the similarity among Step 2 \emph{rule‑level genetic algorithm} to remove repetitive rules.
We measure the similarity using \textbf{minimum neighbor distance} in the embedding space.
Specifically, we first represent all instructions in the embedding space using the \texttt{all-mpnet-base-v2} embedding model. Then calculate the minimum distance from the rules to its nearest neighbors in the embedding space using FAISS.
The minimum neighbor distances of rules in Step 2 after removing repetitions are summarized in Figure \ref{fig:min_distanc}.

However, we believe that deduplication should be performed at the rule level in Step 2, rather than post hoc at the image or puzzle level (e.g., after Step 4). While it may seem appealing to apply a similar embedding-based similarity check to image content using CLIP or SigLIP representations \citep{zhai2023sigmoid}, such visual-level deduplication is fundamentally unreliable for logic-based puzzles.

\begin{figure}[!tb]
    \centering
    \includegraphics[width=\textwidth]{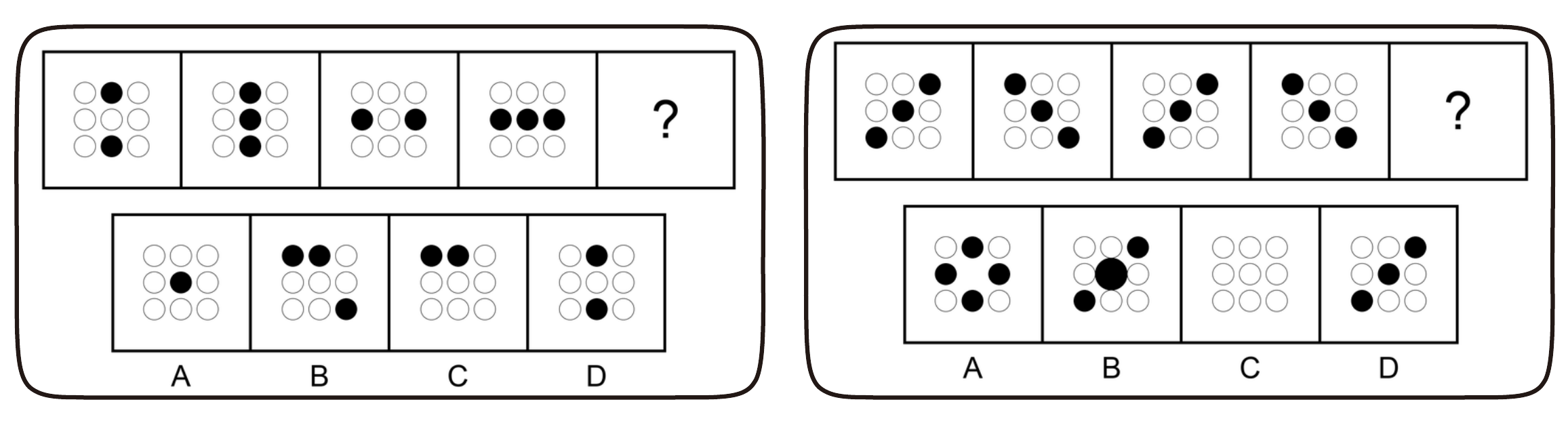}
    \caption{Two puzzles that are visually similar but semantically different.}
    \label{fig:visual_duplicates}
\end{figure}

As illustrated in Figure~\ref{fig:visual_duplicates}, some puzzles—though visually similar due to shared graphical components—encode entirely different reasoning structures and target different answers. In the shown example, two puzzles share nearly identical visual layouts and elements, leading to an extremely small SigLIP-based distance (less than $10^{-4}$), yet they represent distinct logic rules. As a result, any deduplication strategy applied at the visual or puzzle level would mistakenly discard valid, non-redundant instances.

Therefore, semantic-level deduplication must occur immediately after the rule synthesis stage, before these rules are instantiated as images. By enforcing diversity in the rule embedding space at Step 2, we ensure that the downstream generation stages produce puzzles that are both logically and visually distinct.

%% file: tables/classes.tex
\begin{table}[H]
\centering
\caption{Distribution of rule types based on visual pattern and reasoning style. Entries involving \textit{Others} in either dimension are collapsed into the final row.}
\vspace{1em}
\begin{tabular}{l@{\hskip 2em}l@{\hskip 2em}r}
\toprule
\textbf{Visual Pattern} & \textbf{Reasoning Style} & \textbf{Count} \\
\midrule
Horizontal square       & Deductive                & 417 \\
Horizontal square       & Inductive                & 447 \\
Nine-square grid        & Deductive                & 107 \\
Nine-square grid        & Inductive                & 199 \\
Analogy                 & Deductive                & 143 \\
Analogy                 & Inductive                & 118 \\
Two-group               & Inductive                & 464 \\
Others                  & —                        & 499 \\
\bottomrule
\end{tabular}

\label{tab:rule-category-distribution}
\end{table}

%% file: sections/appendix03.tex
\section{Model Selection}
\label{appendix: model selection}

We choose models for each stage of the pipeline based on their availability, modality alignment, reasoning capability, and cost efficiency at the time of experimentation.

\textbf{Step 1.}
When we conducted our experiments, leading models such as GPT-4.1 and Gemini 2.5 \citep{deepmind2025geminipro} had not yet been released or lacked stable APIs. Among the accessible multimodal models, the strongest options were OpenAI's o1 \citep{jaech2024openai} and Anthropic’s Claude-3.7-Sonnet series. We selected Claude-3.5-Haiku and Claude-3.7-Sonnet for their relatively lower costs while maintaining competitive vision-language performance. These models were used to perform image-text verification, explanation rewriting, and rule abstraction for the initial 4K seed questions.

\textbf{Step 2.}
Since this stage only requires text-based reasoning, the candidate space was significantly broader. We selected DeepSeek-V3 due to its strong logical consistency, long-context support, and highly competitive pricing. It was used to power the rule-level genetic algorithm, enabling the expansion from 2.4K seed rules to 60K synthetic rules.

\textbf{Step 3.}
This is the most challenging stage for model selection. Each prompt-response interaction involves generating two Python scripts per rule (correct and distractor), with an average token usage of 2K (non reasoning model) or 12k (reasoning model) per example. The model must not only preserve program logic, but also ensure visual diversity, alignment to abstract rules, and robustness to execution.

\begin{figure}[!tb]
    \centering
    \includegraphics[width=\textwidth]{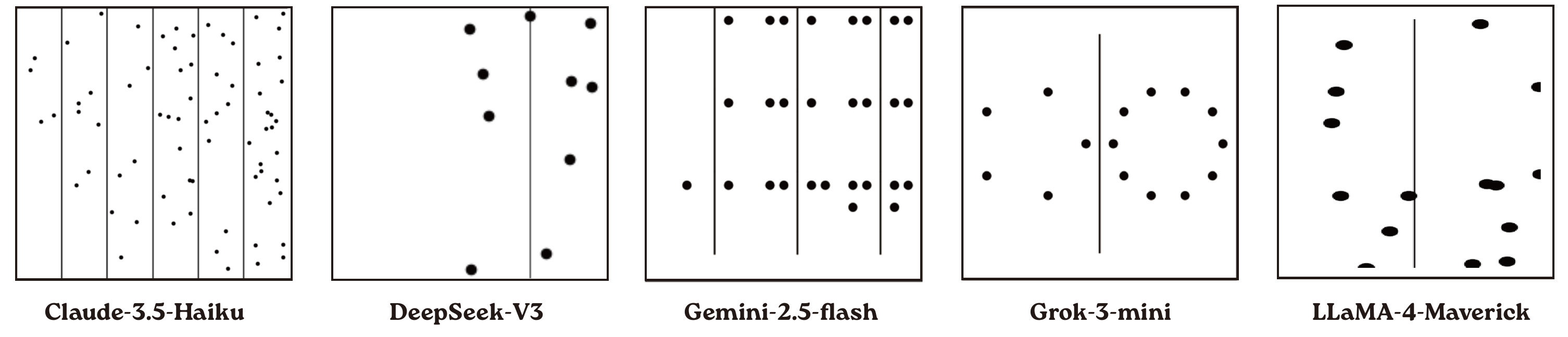}
    \caption{Example comparison of five models on the same rule prompt for image generation.}
    \label{fig:model-example}
\end{figure}

We evaluated several candidate models for this task: Claude-3.5-Haiku, DeepSeek-V3, Gemini-2.5-flash, Grok-3-mini and LLaMA-4-Maverick \citep{meta2025llama4maverick}.
As illustrated in Figure~\ref{fig:model-example}, we show one representative image (out of eight) from each group generated by five different models on the same input prompt. Both Grok-3-mini and Gemini-2.5-flash produce rule-compliant outputs and render images correctly. Among the models we qualitatively compared, Grok-3-mini showed a higher success rate in generating correct and executable scripts, with fewer rendering failures or visual artifacts. Given its reliability and cost-effectiveness, we selected Grok-3-mini as the default model for rule-to-image script generation.

\textbf{Step 4.}
We use GPT-4.1-mini for scoring readability and logical coherence across 110K assembled puzzles. This model offers strong visual-language understanding at minimal cost, making it ideal for large-scale quality control. In parallel, a custom fine-tuned version of Qwen2.5-VL-7B-Instruct is used to estimate model PassRate.

%% file: sections/appendix04.tex
\section{Puzzle Quality Analysis}
\label{appendix: human analysis}

Three experts evaluated a stratified random sample of 1,200 generated puzzles, with 400 puzzles from each difficulty level (easy, medium, and hard). They assessed overall puzzle quality relative to the human-authored seed puzzles, considering logical coherence and complexity. The assessment identified 935 of the 1,200 puzzles (approximately 78\%) as comparable to or better than the seed puzzles. This holistic quality assessment complements the automatic rule-conditioned checks; it does not separately estimate the rates of incorrect, ambiguous, or non-unique-answer puzzles.

%% file: sections/appendix05.tex
\section{Experimental Setups for RL Training}
\label{appendix: experimental setups for RL training}

We conduct reinforcement learning (RL) using the VERL framework~\citep{Sheng_2025} on 8 H100 GPUs with the hyperparameters in Table~\ref{tab: rl hyperparameters}. Reported benchmark comparisons use the step-256 checkpoint for both the 3B and 7B models. The 7B run continued to step 320 to inspect the learning trajectory. We assign binary rewards of 1 for a correct response and 0 for an incorrect response, without additional rewards for formatting. For each question, we append the instruction: ``Let's think step by step and output the final answer within \verb|\\boxed{}|.''

\begin{table}[H]
\small
\centering
\caption{This table shows the hyper-parameters for RL training.}
\vspace{1em}
\begin{tabular}{ll}
\toprule
\textbf{Hyper-parameter} & \textbf{Value} \\ \midrule
Learning Rate & $1 \times 10^{-6}$ \\
Evaluated Checkpoint (Step) & $256$ \\
Extended 7B Run (Steps) & $320$ \\
Number of Devices & $8$ \\
Rollout Batch Size & $64$ \\
PPO Mini Batch Size & $64$ \\
Max Prompt Length & $6000$ \\
Max Response Length & $2048$ \\
KL Coefficient & $0.001$ \\
Rollout Engine & $\textsc{vllm (v0.8.2)}$ \\
Optimizer & \texttt{Adamw} \\
Learning Rate Scheduler & \texttt{cosine} \\
Warmup Ratio & $0.0$ \\ \bottomrule
\end{tabular}
\label{tab: rl hyperparameters}
\end{table}

\section{Experimental Setups for SFT}
\label{appendix: experimental setups for sft}

\paragraph{SFT datasets and construction.}
We construct two SFT datasets from \dataname{}, each containing 10{,}000 final training examples:
\begin{itemize}
\item \textbf{SFT-Random}: randomly select 10{,}000 puzzles and pair each with one response generated by \textsc{Qwen2.5-VL-7B-Instruct}. 
\item \textbf{SFT-Reject}: for each candidate puzzle, sample up to $k{=}5$ responses from \textsc{Qwen2.5-VL-32B-Instruct} and retain the first whose final answer matches the ground truth. Puzzles with no correct response after $5$ attempts are excluded; the final dataset contains 10{,}000 puzzle--response pairs.
\end{itemize}
For response generation during dataset construction, we use near-greedy decoding with temperature $10^{-6}$ and a mild repetition penalty ($1.05$), and require the final answer to be wrapped in \verb|\boxed{}|. Prompts follow the same instruction style used in RL.

\paragraph{Hyper-parameters for SFT.}
We train using LLaMA-Factory \citep{zheng2024llamafactory} with DeepSpeed ZeRO-3 on 8~H100 GPUs, using the hyperparameter detailed in Table~\ref{tab: sft hyperparameters}.

\begin{table}[H]
\small
\centering
\caption{Hyper-parameters for SFT training.}
\vspace{0.8em}
\begin{tabular}{ll}
\toprule
\textbf{Hyper-parameter} & \textbf{Value} \\ \midrule
Effective Batch Size & $128$ (=$8$ per device $\times$ $8$ GPUs $\times$ GA $2$) \\
Per-Device Train Batch Size & $8$ \\
Gradient Accumulation Steps & $2$ \\
Learning Rate & $2 \times 10^{-5}$ \\
Epochs & $3$ \\
LR Scheduler & \texttt{cosine} \\
Warmup Ratio & $0.1$ \\
Optimizer & \texttt{AdamW} \\
Cutoff (Max Seq Length) & $8192$ tokens \\
Chat Template & \texttt{qwen2\_vl} \\
Distributed Config & DeepSpeed ZeRO-3 \\
\bottomrule
\end{tabular}
\label{tab: sft hyperparameters}
\end{table}

\paragraph{Notes.}
We train with full-parameter updates (no LoRA), and keep the same instruction style as RL to standardize evaluation.

\section{Evaluation Details}
\label{appendix: evaluation details}

Each reported training configuration was run once. For all evaluations in Section~\ref{sec: performance analysis}, we use each benchmark's official prompt or evaluation template when available and decode greedily with temperature 0. We do not use additional sampling, self-consistency, hidden prompting, or judge-based correction unless such a step is part of a benchmark's official evaluation protocol. Deterministic decoding does not establish statistical significance. We do not estimate variance from training randomness, checkpoint selection, or data subsampling, so we interpret small transfer gains cautiously.

\paragraph{Answer extraction and scoring.}
For multiple-choice benchmarks, we extract the final option letter from the model output, prioritizing answers wrapped in \verb|\boxed{}| when present and otherwise using the final valid option mention. The extracted option is compared with the benchmark's ground-truth option. For short-answer math or coding-style benchmarks, we follow the benchmark-defined normalization and exact-match rules when available; otherwise, we extract the final answer span from the model's final response and compare it against the official ground truth. This deterministic extraction keeps the evaluation reproducible and avoids introducing an additional model into the scoring pipeline.

\paragraph{\dataname{}-\textsc{Bench}.}
For \dataname{}-\textsc{Bench}, all questions are multiple-choice puzzles. We append the same instruction used in RL training, decode with temperature 0, extract the final answer option from \verb|\boxed{}| or the last valid option letter, and score the prediction against the program-grounded correct option. The full benchmark measures held-out performance within the proposed \dataname{} distribution. To construct the seed-rule-disjoint subset in Table~\ref{tab:seed_disjoint}, we trace all ancestral seed rules of each benchmark puzzle's generating rule and compare them with the union of ancestral seed rules used by the 10K RL training examples. For each of the 71 retained puzzles, its ancestral seed set has an empty intersection with this training ancestry set. This evaluates generalization to puzzles with disjoint ancestral seed rules; it does not imply that the resulting rules are semantically unrelated.

\input{tables/seed_disjoint}

\paragraph{Data and code availability.}
Public resources include the \href{https://huggingface.co/datasets/VisualSphinx/VisualSphinx-V1-Raw}{raw puzzle dataset}, the \href{https://huggingface.co/datasets/VisualSphinx/VisualSphinx-V1-Rules}{rule collection} with \texttt{parents} and \texttt{generation} metadata for tracing ancestry, the \href{https://huggingface.co/datasets/VisualSphinx/VisualSphinx-V1-RL-20K}{RL-20K release}, and the \href{https://huggingface.co/datasets/VisualSphinx/VisualSphinx-V1-Benchmark}{benchmark release}. The \href{https://github.com/VisualSphinx/VisualSphinx}{generation code} is available on GitHub.

\section{Benchmark Details}
\label{appendix: benchmark details}

In this section, we provide detailed descriptions of the benchmarks used to evaluate our models in Section~\ref{sec: performance analysis}.

\paragraph{VisuLogic.} VisuLogic is a visual reasoning benchmark for multimodal large language models that emphasizes vision-centric logic problems and reduces language-only shortcuts. We use it as an external visual-logic benchmark whose task construction differs from the \dataname{} generation pipeline.

\paragraph{RAVEN.} RAVEN is a Raven's Progressive Matrices-style benchmark for relational and analogical visual reasoning. It contains multiple structured layouts, including center, distributed, and paired configurations. We report the average across all seven layouts in Table~\ref{tab:model_comparison}; layout-level results are provided in Table~\ref{tab:raven_details}.

\input{tables/external_logic}

\paragraph{MathVista-testmini.} MathVista-testmini is a widely adopted benchmark consisting of 1,000 visual math problems designed to assess mathematical reasoning in visual contexts. The benchmark evaluates performance across multiple dimensions: \textit{task types}—including Figure Question Answering (FQA), Geometry Problem Solving (GPS), Math Word Problem (MWP), Textbook Question Answering (TQA), and Visual Question Answering (VQA); and \textit{mathematical reasoning types}—including Algebraic Reasoning (ALG), Arithmetic Reasoning (ARI), Geometry Reasoning (GEO), Logical Reasoning (LOG), Numeric Common Sense (NUM), Scientific Reasoning (SCI), and Statistical Reasoning (STA). We report the overall average accuracy across all categories.

\begin{figure}[!tb]
    \centering
    \includegraphics[width=0.5\textwidth]{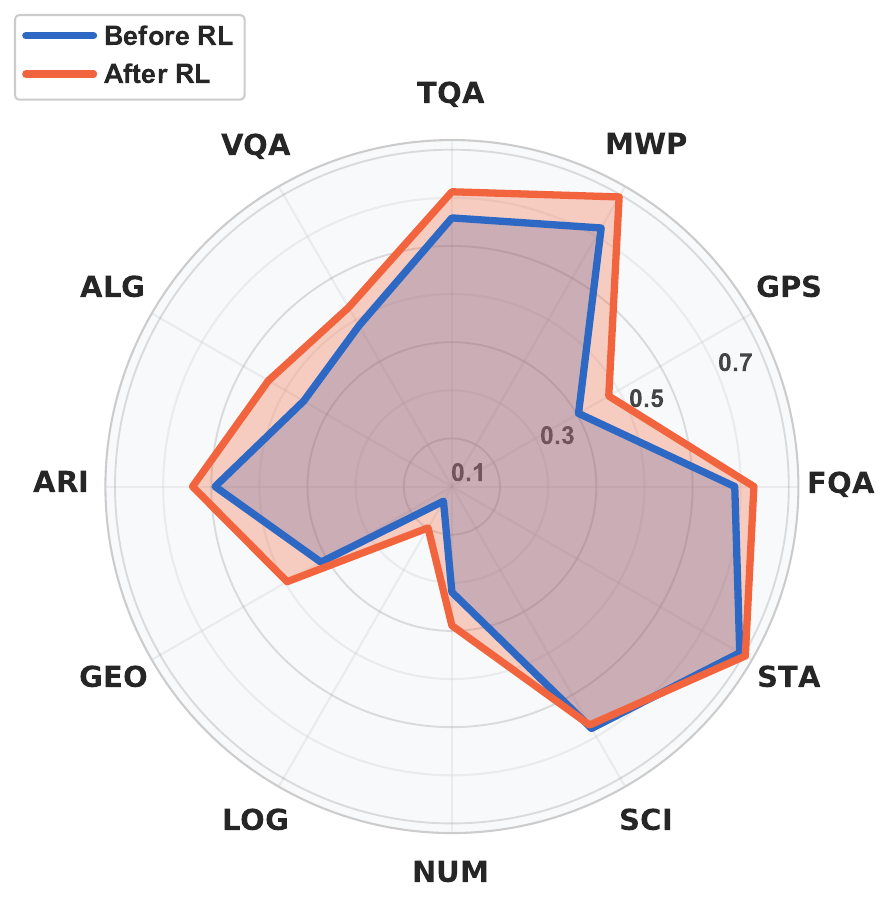}
    \caption{Detailed performance breakdown on MathVista-testmini before and after RL training on \dataname{}. The radar chart shows improvements across multiple reasoning types, with particularly strong gains in logical reasoning (LOG), algebraic reasoning (ALG), and arithmetic reasoning (ARI).}
    \label{fig:radha}
    \vspace{-1em}
\end{figure}

Figure~\ref{fig:radha} shows the MathVista-testmini subcategory breakdown. The overall accuracy increases from 60.0\% to 65.3\%, with notable gains in logical, algebraic, and arithmetic reasoning, demonstrating \dataname{}'s effectiveness in enhancing diverse mathematical reasoning capabilities.

\paragraph{MathVerse-vision.} MathVerse-vision focuses on evaluating vision-dependent mathematical reasoning, where visual information is essential for solving the problems. This benchmark tests the model's ability to extract and reason with mathematical content embedded in images.

\paragraph{We-Math-mini.} We-Math-mini is a curated subset of visual math word problems that emphasizes multi-step reasoning and problem decomposition in visual mathematical contexts.

\paragraph{MMMU-pro standard.} MMMU-pro is a multi-discipline multimodal understanding benchmark covering subjects including science, engineering, humanities, and social sciences. The standard version evaluates general-purpose multimodal reasoning capabilities beyond mathematics.

\paragraph{Algorithm.} The Algorithm benchmark tests algorithmic problem-solving abilities with visual inputs, requiring models to understand visual representations of algorithms, data structures, and computational processes, and generate appropriate solutions or code.

\paragraph{HumanEval-V.} HumanEval-V adapts the classic HumanEval benchmark to include visual context, evaluating whether models can comprehend visual specifications or diagrams and generate corresponding code implementations.

\paragraph{AIME25.} AIME25 contains problems from the 2025 American Invitational Mathematics Examination, a prestigious competition featuring challenging proof-based and computational problems that test advanced mathematical reasoning without visual inputs.

\paragraph{AMC.} The AMC benchmark includes problems from the American Mathematics Competitions (AMC 10 and AMC 12), which assess mathematical problem-solving skills across algebra, geometry, number theory, and combinatorics in a pure text format.

\paragraph{Olympiad.} The Olympiad benchmark features International Mathematical Olympiad-style problems at the highest pre-collegiate level, requiring deep reasoning, creativity, and rigorous proof construction.

\paragraph{MATH-500.} MATH-500 is a curated subset of 500 challenging competition mathematics problems spanning algebra, number theory, counting and probability, geometry, and precalculus, designed to evaluate advanced mathematical reasoning in pure text contexts.

%% file: tables/seed_disjoint.tex
\begin{table}[H]
  \centering
  \caption{Accuracy on the seed-rule-disjoint subset of \dataname{}-\textsc{Bench}. The 71 examples in this subset have underlying seed rules that are entirely disjoint from the seed rules used to generate the 10K RL training set.}
  \vspace{0.5em}
  \label{tab:seed_disjoint}
  \small
  \begin{tabular}{lcc}
    \toprule
    \textbf{Model} & \textbf{Accuracy (\%)} & \textbf{$\Delta$ vs. Instruct} \\
    \midrule
    \textsc{Qwen2.5-VL-3B-Instruct} & 23.94 & -- \\
    \textsc{Qwen2.5-VL-3B-\dataname{}} & \textbf{49.30} & +25.36 \\
    \textsc{Qwen2.5-VL-7B-Instruct} & 28.17 & -- \\
    \textsc{Qwen2.5-VL-7B-\dataname{}} & \textbf{56.34} & +28.17 \\
    \bottomrule
  \end{tabular}
\end{table}

%% file: tables/external_logic.tex
\begin{table}[H]
  \centering
  \caption{Layout-level external visual-logic evaluation on RAVEN. Accuracy is reported in percentage points across seven RAVEN layouts. The average row is reported as the RAVEN result in Table~\ref{tab:model_comparison}.}
  \vspace{0.5em}
  \label{tab:raven_details}
  \small
  \resizebox{\textwidth}{!}{
  \begin{tabular}{lcccc}
    \toprule
    \textbf{Layout} & \textbf{3B Instruct} & \textbf{3B-\dataname{}} & \textbf{7B Instruct} & \textbf{7B-\dataname{}} \\
    \midrule
    \texttt{center\_single} & 17.15 & 24.90 & 28.55 & 33.65 \\
    \texttt{distribute\_four} & 11.75 & 17.00 & 18.90 & 21.80 \\
    \texttt{distribute\_nine} & 11.40 & 13.30 & 15.55 & 16.90 \\
    \texttt{in\_center\_single\_out\_center\_single} & 14.65 & 20.90 & 26.30 & 36.55 \\
    \texttt{in\_distribute\_four\_out\_center\_single} & 13.20 & 22.15 & 23.95 & 33.55 \\
    \texttt{left\_center\_single\_right\_center\_single} & 11.40 & 12.65 & 13.85 & 22.20 \\
    \texttt{up\_center\_single\_down\_center\_single} & 12.55 & 17.20 & 12.80 & 16.55 \\
    \midrule
    \textbf{Average} & 13.16 & \textbf{18.30} & 19.99 & \textbf{25.89} \\
    \bottomrule
  \end{tabular}
  }
\end{table}

%% file: sections/appendix06.tex
\section{Prompt Templates}

\newcommand{\promptanchor}[1]{\phantomsection\refstepcounter{figure}\label{#1}}
\newcommand{\promptcaption}[1]{%
  \begin{center}
    \small Figure~\thefigure: #1
  \end{center}
}

\subsection{Prompt For Step 1}
\label{appendix: Prompt For Step 1}

Figure~\ref{fig:rewirte prompt}-\ref{fig:categorization prompt} demonstrate the prompt template for Step 1 of our pipeline.

\promptanchor{fig:rewirte prompt}
\begin{tcolorbox}[title=Prompt for rewriting explanation and translating into English, promptstyle, breakable]
\lstset{
    basicstyle=\normalfont\sffamily\footnotesize,
    breaklines=true,
    frame=none,
    columns=fullflexible,
}
\begin{CJK*}{UTF8}{gbsn}
\begin{minted}[
    fontsize=\footnotesize,
    breaklines=true,
    breakautoindent=false,
    breakindent=0pt,
    breaksymbolleft=\null,
    breaksymbolindentleft=0pt,
]{text}
You will be translating a Chinese explanation of a vision logic puzzle into English while making certain modifications.

Follow these steps to complete the task:
1. Carefully read and understand the entire Chinese explanation.
2. Translate the explanation from Chinese to English, ensuring that you capture the full meaning and logical relationships in the original text.
3. As you translate, remove any statements that indicate the answer.
4. Remove phrases that mean "如下图所示" ("as shown in the following figure") or any similar references to figures or images. When you remove these phrases, make sure to maintain the original meaning of the statement. For example, if the original text says "如下图所示，有三个圆圈" ("as shown in the following figure, there are three circles"), your translation should simply state "There are three circles."
5. Ensure that you fully understand the logical flow and reasoning of the explanation before finalizing your translation. The goal is to provide a clear, coherent explanation in English that preserves the puzzle's logic without giving away the answer or referring to non-existent figures.
6. Write your final translated and modified explanation in English inside <translated_explanation> tags.

Remember, the key is to provide a faithful translation that maintains the puzzle's logic and difficulty while removing direct answer indicators and references to figures.

For example 1, the text you are going to translate is:
<chinese_explanation>
元素组成不同，且无明显属性规律，考虑数量规律。九宫格第二行出现单独的直线，考虑数直线，但整体直线没有规律，考虑线的细化考法，即线的方向，如下图所示，第一行图一有1组平行线，图二有2组平行线，图三有3组平行线，第二行也满足此规律，第三行前两个图也符合规律，故问号处应有3组平行线，只有D项符合。故正确答案为D。
</chinese_explanation>

Then your output is:
<translated_explanation>
The elements have different compositions with no obvious attribute patterns, so consider quantity patterns. In the second row of the grid, a single line appears. Consider counting lines, but the overall line count lacks a clear pattern. Consider a more refined approach by examining line directions. In the first row, the first image has 1 group of parallel lines, the second image has 2 groups of parallel lines, and the third image has 3 groups of parallel lines. The second row follows a similar pattern, and the first two images in the third row also conform to this pattern. Therefore, the question mark location should follow this progression of parallel line groups.
</translated_explanation>

For example 2, the text you are going to translate is:
<chinese_explanation>
*[omitted for brevity]*
</chinese_explanation>

Then your output is:
<translated_explanation>
*[omitted for brevity]*
</translated_explanation>

Now, please translate the following explanation
<chinese_explanation>
{chinese_explanation_placeholder}
</chinese_explanation>
\end{minted}
\end{CJK*}
\end{tcolorbox}
\promptcaption{Prompt for rewriting and translating}

\promptanchor{fig:verify prompt}
\begin{tcolorbox}[title=Prompt for verifying the question, promptstyle, breakable]
\lstset{
    basicstyle=\normalfont\sffamily\footnotesize,
    breaklines=true,
    frame=none,
    columns=fullflexible,
}
\begin{minted}[
    fontsize=\footnotesize,
    breaklines=true,
    breakautoindent=false,
    breakindent=0pt,
    breaksymbolleft=\null,
    breaksymbolindentleft=0pt,
]{text}
## Task Description
You will be given an image of a vision logic puzzle. Your task is to analyze the image, understand the puzzle's logic, and provide a detailed explanation of how to solve it.

## Input
<image>
<!--SPLIT-->
</image>

<question>
{{question}}
</question>

<options>
{{options}}
</options>

<hint>
{{hint}}
</hint>

## Instructions
1. Carefully examine the image, paying attention to all visual elements, patterns, and relationships between objects.  
2. Consider the question (and hint) to guide your analysis.  
3. Identify the underlying logic or rule that governs the puzzle.  
4. Apply this logic to determine the correct answer.  
5. Explain your reasoning step by step, referencing specific elements in the image.  
6. **Answer in English.**

## Output
<reasoning>
Your reasoning.
</reasoning>

<answer>
Your answer (select only from: A, B, C, D, or Problematic)
</answer>
\end{minted}
\end{tcolorbox}
\promptcaption{Prompt for verifying the question}

\promptanchor{fig:abstract prompt}
\begin{tcolorbox}[title=Prompt for abstracting the puzzle into pullet points, promptstyle, breakable]
\lstset{
    basicstyle=\normalfont\sffamily\footnotesize,
    breaklines=true,
    frame=none,
    columns=fullflexible,
}
\begin{minted}[
    fontsize=\footnotesize,
    breaklines=true,
    breakautoindent=false,
    breakindent=0pt,
    breaksymbolleft=\null,
    breaksymbolindentleft=0pt,
]{text}
## Task Description
You are an expert in analyzing and summarizing complex vision logic puzzles. Your task is to extract and present the key regularities and patterns from a given puzzle and its explanation. The goal is to create a concise list of regularities that captures the essential elements of the puzzle and its solution, which could be used to generate similar puzzles.
Now, follow these steps to analyze the puzzle and extract key regularities:
1. **Analyze** the puzzle and its explanation thoroughly.  
2. **Create** a detailed breakdown of the puzzle inside `<detailed_analysis>` tags …  
3. **Present** your analysis within `<puzzle_breakdown>` tags …  
4. **Create** a list of key regularities within `<key_points>` tags …  
5. **Review & refine** your list …  

### Example
Examine the image of the puzzle:  
<image><!--EXAMPLE_SPLIT--></image>
Read the question:  
<puzzle_question>  
From the given four options, select the most suitable one to fill in the question mark to present a certain regularity:  
</puzzle_question>
Review the options:  
<puzzle_options>
*[omitted for brevity]*
</puzzle_options>
Read the explanation:  
<puzzle_explanation>  
The elements have similar compositions … *[omitted for brevity]*  
</puzzle_explanation>
Note the answer:  
<puzzle_answer>  
D  
</puzzle_answer>
The model answer is:
<detailed_analysis></detailed_analysis>
<puzzle_breakdown></puzzle_breakdown>
<key_points>
- Inner shapes become outer shapes in the subsequent panel  
- Shapes alternate between curved and straight-edged forms  
- Each panel contains exactly two nested geometric shapes  
- Multiple pattern layers must be satisfied simultaneously  
- Shape orientation varies while maintaining structural patterns  
</key_points>

## Input
Examine the image of the puzzle:  
<image>
<!--PUZZLE_SPLIT-->
</image>
Read the question:  
*[omitted for brevity]*

## Output
Your final output **must** follow this exact structure:
<detailed_analysis>
[Your detailed analysis here]
</detailed_analysis>
<puzzle_breakdown>
[Your structured breakdown here]
</puzzle_breakdown>
<key_points>
- [Regularity 1]  
- [Regularity 2]  
- … (max 5 items, each less than 30 words)  
</key_points>
\end{minted}
\end{tcolorbox}
\promptcaption{Prompt for abstracting the puzzle into pullet points}

\promptanchor{fig:categorization prompt}
\begin{tcolorbox}[title=Prompt for categorization, promptstyle, breakable]
\lstset{
    basicstyle=\normalfont\sffamily\footnotesize,
    breaklines=true,
    frame=none,
    columns=fullflexible,
}
\begin{minted}[
    fontsize=\footnotesize,
    breaklines=true,
    breakautoindent=false,
    breakindent=0pt,
    breaksymbolleft=\null,
    breaksymbolindentleft=0pt,
]{text}
## Task Description
You are an expert system designed to analyze and classify complex visual logic puzzles.  
Your task is to examine the given puzzle components and determine **one** question‑type tag and **one** knowledge‑point tag from the predefined lists.

### Question‑type tags
- **Nine-square grid** A 3 × 3 grid with one cell containing a question mark.  
- **Horizontal square** A row of 4–6 squares with one containing a question mark.  
- **Two-group** Two groups of three squares; one square of one group has a question mark.  
- **Two set of number** Images with numbers that must be divided into two sets.  
- **Others** Doesn’t fit any category above.

### Knowledge‑point tags
- **Correlated** Each image is directly related to the previous one.  
- **Summarize** All images share an overall rule but adjacent images may not correlate.  
- **Others** Doesn’t fit any category above.

## Analysis instructions
1. Examine every puzzle component.  
2. Focus on structure and relationships.  
3. Weigh arguments **for and against** each tag.  
4. Provide detailed reasoning inside `<puzzle_breakdown>` before giving tags.

## Input

Here is the puzzle you need to analyze:

<puzzle_image>
<!--PUZZLE_SPLIT-->
</puzzle_image>

<puzzle_question>
{{ prompt }}
</puzzle_question>

<puzzle_options>
{{ options_block }}
</puzzle_options>

<puzzle_explanation>
{{ explanation }}
</puzzle_explanation>

<puzzle_answer>
{{ correct_answer }}
</puzzle_answer>

## Output (required structure)

<puzzle_breakdown>
[Your detailed reasoning here]
</puzzle_breakdown>

<question_type>
[Selected question‑type tag]
</question_type>

<knowledge_point>
[Selected knowledge‑point tag]
</knowledge_point>
\end{minted}
\end{tcolorbox}
\promptcaption{Prompt for categorization}

\subsection{Prompt For Step 2 - Rule Expansion}
\label{appendix: Prompt For Rule Expansion}

Figure~\ref{fig:crossover prompt}-\ref{fig:mutation prompt} demonstrate the prompt template for Step 2 rule-level genetic algorithm.

\promptanchor{fig:crossover prompt}
\begin{tcolorbox}[title=Prompt for crossover, promptstyle, breakable]
\lstset{
    basicstyle=\normalfont\sffamily\footnotesize,
    breaklines=true,
    frame=none,
    columns=fullflexible,
}
\begin{minted}[
    fontsize=\footnotesize,
    breaklines=true,
    breakautoindent=false,
    breakindent=0pt,
    breaksymbolleft=\null,
    breaksymbolindentleft=0pt,
]{text}
## Task Description
You are an expert in analyzing and synthesizing complex vision logic puzzles. Your task is to crossover and combine two sets of puzzle regularities to create a novel, coherent rule set that could use to generate new puzzles. Rather than mechanically combining individual points, you need to understand the underlying logic and context of both rule sets to create meaningful hybrid patterns that maintain internal consistency. 

Please follow these steps:

1. Analyze both rule sets:
   - List out key elements from each rule set
   - Identify core principles, compatibility, and potential synergies
   - Look for opportunities to create emergent patterns through combination
   - Identify potential conflicts and propose resolutions

2. Wrap your detailed analysis in <comparative_analysis> tags:
   - Compare and contrast the rule sets
   - Identify complementary regularities and explore potential synergies
   - Consider application to puzzle generation
   - Analyze and propose resolutions for potential conflicts

3. Wrap your cross-over synthesis in <synthesis> tags:
   - Explain how you've integrated elements from both rule sets
   - Describe how the new combined regularities work together
   - Illustrate potential puzzle scenarios
   - Explain the coherence and unity of your cross-over
   - Discuss how it maintains core insights and ensures contextual continuity

4. Wrap your final cross-over rule set in <crossover_rules> tags:
   - Each rule should be less than 30 words and incorporate elements from both source rule sets
   - Aim for 4-6 cohesive items that flow logically and create a complete puzzle system
   - Ensure the rules could plausibly generate square-format continuous image puzzles
   - Check for contextual integrity, preservation of fundamental insights, and logical progression

## Input
Here are the two sets of puzzle regularities:

<first_rule_set>
{FIRST_RULE_SET}
</first_rule_set>

<second_rule_set>
{SECOND_RULE_SET}
</second_rule_set>

## Output
Your final output should follow this structure:

<comparative_analysis>
[Your detailed comparative analysis]
</comparative_analysis>

<synthesis>
[Your explanation of the integrated elements and how they work together]
</synthesis>

<crossover_rules>
- [Cross-over Rule 1]
- [Cross-over Rule 2]
- [Cross-over Rule 3]
- [Cross-over Rule 4]
- [Cross-over Rule 5]
- [Cross-over Rule 6] (optional)
</crossover_rules>

Remember, your cross-over should be a thoughtful integration that preserves context and creates emergent properties, not a mechanical merging of the original rules.
\end{minted}
\end{tcolorbox}
\promptcaption{Prompt for crossover}

\promptanchor{fig:mutation prompt}
\begin{tcolorbox}[title=Prompt for mutation, promptstyle, breakable]
\lstset{
    basicstyle=\normalfont\sffamily\footnotesize,
    breaklines=true,
    frame=none,
    columns=fullflexible,
}
\begin{minted}[
    fontsize=\footnotesize,
    breaklines=true,
    breakautoindent=false,
    breakindent=0pt,
    breaksymbolleft=\null,
    breaksymbolindentleft=0pt,
]{text}
## Task Description
You are an expert in analyzing and evolving complex vision logic puzzles. Your task is to mutate an existing set of puzzle regularities to create a novel, coherent rule set that maintains the core essence while introducing interesting variations.

Please follow these steps:

1. Analyze the existing rule set:
   - Identify the core principles and key elements
   - Understand the underlying logic and structure

2. Wrap your analysis in <analysis> tags:
   - Describe key insights about the rule structure
   - Identify opportunities for creative mutation
   - Explain which aspects should be preserved vs. modified

3. Create a mutated version with controlled changes:
   - Introduce meaningful variations to one point or two points of the rules
   - Ensure the mutations maintain logical coherence
   - Preserve the core puzzle structure and logic
   - Aim for a mutation that could generate novel but solvable puzzles

4. Wrap your final mutated rule set in <mutated_rules> tags:
   - Each rule should be less than 30 words
   - Include 4-6 rules total (similar to the input set)
   - Ensure the mutated rules maintain logical consistency
   - The mutation should be meaningful but not completely transform the puzzle type

## Input
Here is the set of puzzle regularities to mutate:

<rule_set>
{RULE_SET}
</rule_set>

## Output
Your final output should follow this structure:

<analysis>
[Your analysis of the existing rules]
</analysis>

<mutated_rules>
- [Mutated Rule 1]
- [Mutated Rule 2]
- [Mutated Rule 3]
- [Mutated Rule 4]
- [Mutated Rule 5]
- [Mutated Rule 6] (optional)
</mutated_rules>

Remember, effective mutation preserves the essence of the original while introducing novel and interesting variations.
\end{minted}
\end{tcolorbox}
\promptcaption{Prompt for mutation}

\subsection{Prompt For Step 2 - Scoring}
\label{appendix: Prompt For Scoring}

Figure~\ref{fig:scoring prompt} demonstrates the prompt template for scoring synthetic rules.

\promptanchor{fig:scoring prompt}
\begin{tcolorbox}[title=Prompt for scoring synthetic rules, promptstyle, breakable]
\lstset{
    basicstyle=\normalfont\sffamily\footnotesize,
    breaklines=true,
    frame=none,
    columns=fullflexible,
}
\begin{minted}[
    fontsize=\footnotesize,
    breaklines=true,
    breakautoindent=false,
    breakindent=0pt,
    breaksymbolleft=\null,
    breaksymbolindentleft=0pt,
]{text}
## Task Description
You are an expert evaluator tasked with assessing a rule for a visual logical reasoning question. This rule is designed to guide a large language model in writing Python code to generate a sequence of images that correspond to the rule. Your evaluation will focus on three key criteria: format, content quality, and feasibility.

Before providing your final evaluation, please wrap your detailed analysis for each criterion in <detailed_analysis> tags. This will ensure a thorough and transparent assessment.

Evaluation Criteria and Rubric:
1. Format (1-5 points)
   1: Incomplete, missing multiple points, contains unrelated content
   2: Incomplete, missing 1-2 points, may contain some unrelated content
   3: Complete set of points, but may have some unrelated content
   4: Complete set of 5-6 points, minimal unrelated content
   5: Perfect format with 5-6 points and no unrelated content
2. Content Quality (1-5 points)
   1: Inconsistent, contradictory, does not describe core concept
   2: Some inconsistencies or contradictions, poorly describes core concept
   3: Mostly consistent, few contradictions, adequately describes core concept
   4: Consistent, no major contradictions, effectively describes core concept
   5: Highly consistent, no contradictions at all, perfectly describes core concept
3. Feasibility (1-5 points)
   1: Vague, unclear instructions, not suitable for code generation
   2: Some clear instructions, but many gaps for code generation
   3: Mostly clear instructions, some gaps for code generation
   4: Clear instructions, minor gaps for code generation
   5: Perfectly clear and specific instructions for easy code generation

## Input
Here is the rule you will evaluate:
<rule>
{RULE}
</rule>

## Output
Please evaluate the rule using the following structure:
<detailed_analysis>
Format Analysis:
- List key points of the rule relevant to format
- Analyze each point in relation to the format scoring rubric
- Provide a preliminary format score based on the analysis
Content Quality Analysis:
- List key points of the rule relevant to content quality
- Analyze each point in relation to the content quality scoring rubric
- Provide a preliminary content quality score based on the analysis
Feasibility Analysis:
- List key points of the rule relevant to feasibility
- Analyze each point in relation to the feasibility scoring rubric
- Provide a preliminary feasibility score based on the analysis
</detailed_analysis>

## Final Score
<format_score>
[Score] (1~5)
</format_score>
<content_quality>
[Score] (1~5)
</content_quality>
<feasibility>
[Score] (1~5)
</feasibility>
\end{minted}
\end{tcolorbox}
\promptcaption{Prompt for scoring synthetic rules}

\subsection{Prompt For Step 3}
\label{appendix: Prompt For Step 3}

Figure~\ref{fig:style 1 prompt}-\ref{fig:style 3 prompt} demonstrate the prompt template for three rule-to-image rendering styles.

\promptanchor{fig:style 1 prompt}
\begin{tcolorbox}[title=Prompt for first rule-to-image rendering style, promptstyle, breakable]
\lstset{
    basicstyle=\normalfont\sffamily\footnotesize,
    breaklines=true,
    frame=none,
    columns=fullflexible,
}
\begin{minted}[
    fontsize=\footnotesize,
    breaklines=true,
    breakautoindent=false,
    breakindent=0pt,
    breaksymbolleft=\null,
    breaksymbolindentleft=0pt,
]{text}
## Task Description
As a Python imaging expert, create two Python scripts: one generating 5 continuous images that follow a set of rules, and another generating 3 images that intentionally break those rules. Infer a reasonable implementation according to rules (e.g., based on patterns, shapes, or mathematical properties) and document your reasoning.

## Input
Let's review the rules for your images:

<rules>
{{RULES}}
</rules>

## Guidelines
- **Libraries**: Use matplotlib and numpy.
- **Style**: Black-and-white images, no text.
- **Output Paths**: 
  - Correct images: ./output_correct/
  - Incorrect images: ./output_incorrect/
- **Image Counts**: 5 correct, 3 incorrect.
- **Size**: Adjust the size of elements and images to avoid confusion and unnecessary overlap.
- Optionally, sometimes you can use your continuous output images to express some rules that are related between images instead of expressing all in a single one image.
  
## Deliverables
Provide two complete, runnable Python scripts:
1. **Correct Script**: Generates 5 continuous images complying the rules.
2. **Incorrect Script**: Generates 3 images, each breaking a different rule.

For each script:
- Include all imports.
- Your code should well-documented and add concise comments explaining rule compliance or violation.
- Use relative paths ./output_correct and ./output_incorrect.

## Output
Generate two Python scripts as requested. Present them *exactly* in the following format, with only the raw code inside the tags:

<correct_script>
*Your complete Python script for generating 5 correct images*
</correct_script>

<incorrect_script>
*Your complete Python script for generating 3 incorrect images*
</incorrect_script>

Remember, two scripts should be complete and directly runnable.
\end{minted}
\end{tcolorbox}
\promptcaption{Prompt for first rule-to-image rendering style}

\promptanchor{fig:style 2 prompt}
\begin{tcolorbox}[title=Prompt for second rule-to-image rendering style, promptstyle, breakable]
\lstset{
    basicstyle=\normalfont\sffamily\footnotesize,
    breaklines=true,
    frame=none,
    columns=fullflexible,
}
\begin{minted}[
    fontsize=\footnotesize,
    breaklines=true,
    breakautoindent=false,
    breakindent=0pt,
    breaksymbolleft=\null,
    breaksymbolindentleft=0pt,
]{text}
## Task Description
As a Python imaging expert, create two Python scripts: one generating 5 continuous images that follow a set of rules, and another generating 3 images that intentionally break those rules. Infer a reasonable implementation according to rules (e.g., based on patterns, shapes, or mathematical properties) and document your reasoning.

## Input
Let's review the rules for your images:

<rules>
{{RULES}}
</rules>

## Guidelines
- **Libraries**: Use PIL and numpy.
- **Style**: Black-and-white images, no text.
- **Output Paths**: 
  - Correct images: ./output_correct/
  - Incorrect images: ./output_incorrect/
- **Image Counts**: 5 correct, 3 incorrect.
- **Size**: Adjust the size of elements and images to avoid confusion and unnecessary overlap.
- Optionally, sometimes you can use your continuous output images to express some rules that are related between images instead of expressing all in a single one image.
  
## Deliverables
Provide two complete, runnable Python scripts:
1. **Correct Script**: Generates 5 continuous images complying the rules.
2. **Incorrect Script**: Generates 3 images, each breaking a different rule.

For each script:
- Include all imports.
- Your code should well-documented and add concise comments explaining rule compliance or violation.
- Use relative paths ./output_correct and ./output_incorrect.

## Output
Generate two Python scripts as requested. Present them *exactly* in the following format, with only the raw code inside the tags:

<correct_script>
*Your complete Python script for generating 5 correct images*
</correct_script>

<incorrect_script>
*Your complete Python script for generating 3 incorrect images*
</incorrect_script>

Remember, two scripts should be complete and directly runnable.
\end{minted}
\end{tcolorbox}
\promptcaption{Prompt for second rule-to-image rendering style}

\promptanchor{fig:style 3 prompt}
\begin{tcolorbox}[title=Prompt for third rule-to-image rendering style, promptstyle, breakable]
\lstset{
    basicstyle=\normalfont\sffamily\footnotesize,
    breaklines=true,
    frame=none,
    columns=fullflexible,
}
\begin{minted}[
    fontsize=\footnotesize,
    breaklines=true,
    breakautoindent=false,
    breakindent=0pt,
    breaksymbolleft=\null,
    breaksymbolindentleft=0pt,
]{text}
## Task Description
As a Python imaging expert, create two Python scripts: one generating 5 continuous images that follow a set of rules, and another generating 3 images that intentionally break those rules. Infer a reasonable implementation according to rules (e.g., based on patterns, shapes, or mathematical properties) and document your reasoning.

## Input
Let's review the rules for your images:

<rules>
{{RULES}}
</rules>

## Guidelines
- **Libraries**: No restrictions.
- **Style**: Your color matching should be very reasonable and beautiful. No text on images.
- **Output Paths**: 
  - Correct images: ./output_correct/
  - Incorrect images: ./output_incorrect/
- **Image Counts**: 5 correct, 3 incorrect.
- **Size**: Adjust the size of elements and images to avoid confusion and unnecessary overlap.
- Optionally, sometimes you can use your continuous output images to express some rules that are related between images instead of expressing all in a single one image.
  
## Deliverables
Provide two complete, runnable Python scripts:
1. **Correct Script**: Generates 5 continuous images complying the rules.
2. **Incorrect Script**: Generates 3 images, each breaking a different rule.

For each script:
- Include all imports.
- Your code should well-documented and add concise comments explaining rule compliance or violation.
- Use relative paths ./output_correct and ./output_incorrect.

## Output
Generate two Python scripts as requested. Present them *exactly* in the following format, with only the raw code inside the tags:

<correct_script>
*Your complete Python script for generating 5 correct images*
</correct_script>

<incorrect_script>
*Your complete Python script for generating 3 incorrect images*
</incorrect_script>

Remember, two scripts should be complete and directly runnable.

\end{minted}
\end{tcolorbox}
\promptcaption{Prompt for third rule-to-image rendering style}

\subsection{Prompt for Annotation}
\label{appendix: Prompt for Annotation}

Figure~\ref{fig:puzzles scoring prompt} demonstrates the prompt template for scoring assembled puzzles.

\promptanchor{fig:puzzles scoring prompt}
\begin{tcolorbox}[title=Prompt for scoring assembled puzzles, promptstyle, breakable]
\lstset{
    basicstyle=\normalfont\sffamily\footnotesize,
    breaklines=true,
    frame=none,
    columns=fullflexible,
}
\begin{minted}[
    fontsize=\footnotesize,
    breaklines=true,
    breakautoindent=false,
    breakindent=0pt,
    breaksymbolleft=\null,
    breaksymbolindentleft=0pt,
]{text}
## Task Description
You are an expert evaluator tasked with assessing a vision puzzle consisting of 8 sub-pictures. Your goals in this prompt are:
1. **Question Reasonableness Evaluation:** Evaluate whether the provided question and answer align with the rules and are unambiguous.  
2. **Overall Visual Readability Evaluation:** Assess the clarity and readability of the entire puzzle image.

## Input
<image>
<!--SPLIT-->
</image>

<question>
{{question}}
</question>

<answer>
{{answer}}
</answer>

<rules>
{{rules}}
</rules>

## Instructions

1. **Question Reasonableness Evaluation**  
   Use the provided rules and answer to verify the reasonableness of the question. Score this on a scale of 1–5, where 5 is the most reasonable. Consider these factors:  
   - Does the answer align perfectly with the rule described?  
   - Is there any ambiguity in the question or answer?  
   - Is the provided answer the only correct solution?  
   **Score Criteria:**  
   - **5**: Perfect alignment, no ambiguity, single correct solution.  
   - **4**: Strong alignment, minor ambiguity, likely single solution.  
   - **3**: Reasonable alignment, some ambiguity or alternative interpretations.  
   - **2**: Weak alignment, ambiguous, multiple plausible answers.  
   - **1**: Poor alignment, high ambiguity, answer does not follow rules.  

2. **Overall Visual Readability Evaluation**  
   Assess the overall visual readability of the puzzle image. Score this on a scale of 1–5.  
   **Score Criteria:**  
   - **5**: Perfectly readable with no issues.  
   - **4**: Minor readability issues, but still easy to understand.  
   - **3**: Moderate readability issues that may impact understanding.  
   - **2**: Significant readability issues that hinder understanding.  
   - **1**: Severe readability issues that make the puzzle nearly impossible to understand.  

## Output
<reasonableness_evaluation>
[Your detailed justification here]
Score: [1~5]
</reasonableness_evaluation>

<readability_evaluation>
[Your detailed justification here]
Score: [1~5]
</readability_evaluation>

<final_scores>
Reasonableness: [1~5]
Readability: [1~5]
</final_scores>
\end{minted}
\end{tcolorbox}
\promptcaption{Prompt for scoring assembled puzzles}

%% file: sample.bbl
\begin{thebibliography}{67}
\providecommand{\natexlab}[1]{#1}
\providecommand{\url}[1]{\texttt{#1}}
\expandafter\ifx\csname urlstyle\endcsname\relax
  \providecommand{\doi}[1]{doi: #1}\else
  \providecommand{\doi}{doi: \begingroup \urlstyle{rm}\Url}\fi

\bibitem[Alam et~al.(2025)Alam, Aggarwal, Chae, and Rastogi]{alam2025sphinx}
Md~Tanvirul Alam, Saksham Aggarwal, Justin~Yang Chae, and Nidhi Rastogi.
\newblock Sphinx: A synthetic environment for visual perception and reasoning.
\newblock \emph{arXiv preprint arXiv:2511.20814}, 2025.
\newblock URL \url{https://arxiv.org/abs/2511.20814}.

\bibitem[Anthropic(2024)]{claude3}
Anthropic.
\newblock Introducing the next generation of claude.
\newblock \url{https://www.anthropic.com/news/claude-3-family}, 2024.

\bibitem[Carter(2024)]{textocr-gpt4v}
Jimmy Carter.
\newblock Textocr-gpt4v.
\newblock \url{https://huggingface.co/datasets/jimmycarter/textocr-gpt4v}, 2024.

\bibitem[Chen et~al.(2024{\natexlab{a}})Chen, Chen, Zhang, Chen, Wu, Zhang, Chen, Li, Wan, and Wang]{chen2024allava}
Guiming~Hardy Chen, Shunian Chen, Ruifei Zhang, Junying Chen, Xiangbo Wu, Zhiyi Zhang, Zhihong Chen, Jianquan Li, Xiang Wan, and Benyou Wang.
\newblock Allava: Harnessing gpt4v-synthesized data for lite vision-language models.
\newblock \emph{arXiv preprint arXiv:2402.11684}, 2024{\natexlab{a}}.

\bibitem[Chen et~al.(2023)Chen, Zhang, Ren, Zhao, Cai, Wang, Wang, Liu, and Chang]{chen2023towards}
Liang Chen, Yichi Zhang, Shuhuai Ren, Haozhe Zhao, Zefan Cai, Yuchi Wang, Peiyi Wang, Tianyu Liu, and Baobao Chang.
\newblock Towards end-to-end embodied decision making via multi-modal large language model: Explorations with gpt4-vision and beyond.
\newblock \emph{arXiv preprint arXiv:2310.02071}, 2023.

\bibitem[Chen et~al.(2024{\natexlab{b}})Chen, Li, Dong, Zhang, He, Wang, Zhao, and Lin]{chen2024sharegpt4v}
Lin Chen, Jinsong Li, Xiaoyi Dong, Pan Zhang, Conghui He, Jiaqi Wang, Feng Zhao, and Dahua Lin.
\newblock Sharegpt4v: Improving large multi-modal models with better captions.
\newblock In \emph{European Conference on Computer Vision}, pages 370--387. Springer, 2024{\natexlab{b}}.

\bibitem[Chen et~al.(2021)Chen, Tworek, Jun, Yuan, Pinto, Kaplan, Edwards, Burda, Joseph, Brockman, et~al.]{chen2021evaluating}
Mark Chen, Jerry Tworek, Heewoo Jun, Qiming Yuan, Henrique Ponde De~Oliveira Pinto, Jared Kaplan, Harri Edwards, Yuri Burda, Nicholas Joseph, Greg Brockman, et~al.
\newblock Evaluating large language models trained on code.
\newblock \emph{arXiv preprint arXiv:2107.03374}, 2021.

\bibitem[Chen et~al.(2025)Chen, Qin, Liu, Peng, Guan, Wang, Hu, Zhou, Gao, and Che]{chen2025towards}
Qiguang Chen, Libo Qin, Jinhao Liu, Dengyun Peng, Jiannan Guan, Peng Wang, Mengkang Hu, Yuhang Zhou, Te~Gao, and Wanxiang Che.
\newblock Towards reasoning era: A survey of long chain-of-thought for reasoning large language models.
\newblock \emph{arXiv preprint arXiv:2503.09567}, 2025.

\bibitem[Cobbe et~al.(2021)Cobbe, Kosaraju, Bavarian, Chen, Jun, Kaiser, Plappert, Tworek, Hilton, Nakano, et~al.]{cobbe2021training}
Karl Cobbe, Vineet Kosaraju, Mohammad Bavarian, Mark Chen, Heewoo Jun, Lukasz Kaiser, Matthias Plappert, Jerry Tworek, Jacob Hilton, Reiichiro Nakano, et~al.
\newblock Training verifiers to solve math word problems.
\newblock \emph{arXiv preprint arXiv:2110.14168}, 2021.

\bibitem[Deng et~al.(2024)Deng, Yang, Ding, Liu, Li, Qi, and Ngai]{deng2024can}
Weipeng Deng, Jihan Yang, Runyu Ding, Jiahui Liu, Yijiang Li, Xiaojuan Qi, and Edith Ngai.
\newblock Can 3d vision-language models truly understand natural language?
\newblock \emph{arXiv preprint arXiv:2403.14760}, 2024.

\bibitem[Douze et~al.(2024)Douze, Guzhva, Deng, Johnson, Szilvasy, Mazaré, Lomeli, Hosseini, and Jégou]{douze2024faiss}
Matthijs Douze, Alexandr Guzhva, Chengqi Deng, Jeff Johnson, Gergely Szilvasy, Pierre-Emmanuel Mazaré, Maria Lomeli, Lucas Hosseini, and Hervé Jégou.
\newblock The faiss library.
\newblock \emph{arXiv preprint arXiv:2401.08281}, 2024.

\bibitem[Fang et~al.(2024)Fang, Zhu, Lu, Wang, Molchanov, Kautz, Cho, Pavone, Han, and Yin]{fang2024vila}
Yunhao Fang, Ligeng Zhu, Yao Lu, Yan Wang, Pavlo Molchanov, Jan Kautz, Jang~Hyun Cho, Marco Pavone, Song Han, and Hongxu Yin.
\newblock Vila$^{2}$: Vila augmented vila.
\newblock \emph{arXiv preprint arXiv:2407.17453}, 2024.

\bibitem[Feng et~al.(2023)Feng, Luo, Wang, Tang, Yang, Shao, Mguni, Du, and Wang]{feng2023chessgpt}
Xidong Feng, Yicheng Luo, Ziyan Wang, Hongrui Tang, Mengyue Yang, Kun Shao, David~Henry Mguni, Yali Du, and Jun Wang.
\newblock Chess{GPT}: Bridging policy learning and language modeling.
\newblock In \emph{Thirty-seventh Conference on Neural Information Processing Systems Datasets and Benchmarks Track}, 2023.
\newblock URL \url{https://openreview.net/forum?id=pvdm4B6JMK}.

\bibitem[{Google DeepMind}(2025)]{deepmind2025geminipro}
{Google DeepMind}.
\newblock {Gemini Pro}.
\newblock \url{https://deepmind.google/technologies/gemini/pro/}, March 2025.
\newblock Accessed: 2025-04-03.

\bibitem[Hadmi$^1$ et~al.(2012)Hadmi$^1$, Puech$^1$, and Said]{hadmi12012perceptual}
Azhar Hadmi$^1$, William Puech$^1$, and Brahim Ait~Es Said.
\newblock Perceptual image hashing.
\newblock \emph{Watermarking: Volume 2}, page~17, 2012.

\bibitem[He et~al.(2024)He, Luo, Bai, Hu, Thai, Shen, Hu, Han, Huang, Zhang, et~al.]{he2024olympiadbench}
Chaoqun He, Renjie Luo, Yuzhuo Bai, Shengding Hu, Zhen~Leng Thai, Junhao Shen, Jinyi Hu, Xu~Han, Yujie Huang, Yuxiang Zhang, et~al.
\newblock Olympiadbench: A challenging benchmark for promoting agi with olympiad-level bilingual multimodal scientific problems.
\newblock \emph{arXiv preprint arXiv:2402.14008}, 2024.

\bibitem[Hendrycks et~al.(2021)Hendrycks, Burns, Basart, Zou, Mazeika, Song, and Steinhardt]{hendrycks2021measuring}
Dan Hendrycks, Collin Burns, Steven Basart, Andy Zou, Mantas Mazeika, Dawn Song, and Jacob Steinhardt.
\newblock Measuring massive multitask language understanding.
\newblock In \emph{International Conference on Learning Representations}, 2021.
\newblock URL \url{https://openreview.net/forum?id=d7KBjmI3GmQ}.

\bibitem[Jaech et~al.(2024)Jaech, Kalai, Lerer, Richardson, El-Kishky, Low, Helyar, Madry, Beutel, Carney, et~al.]{jaech2024openai}
Aaron Jaech, Adam Kalai, Adam Lerer, Adam Richardson, Ahmed El-Kishky, Aiden Low, Alec Helyar, Aleksander Madry, Alex Beutel, Alex Carney, et~al.
\newblock Openai o1 system card.
\newblock \emph{arXiv preprint arXiv:2412.16720}, 2024.

\bibitem[Keahey et~al.(2020)Keahey, Anderson, Zhen, Riteau, Ruth, Stanzione, Cevik, Colleran, Gunawi, Hammock, Mambretti, Barnes, Halbah, Rocha, and Stubbs]{keahey2020lessons}
Kate Keahey, Jason Anderson, Zhuo Zhen, Pierre Riteau, Paul Ruth, Dan Stanzione, Mert Cevik, Jacob Colleran, Haryadi~S. Gunawi, Cody Hammock, Joe Mambretti, Alexander Barnes, Fran{\c c}ois Halbah, Alex Rocha, and Joe Stubbs.
\newblock Lessons learned from the chameleon testbed.
\newblock In \emph{2020 USENIX Annual Technical Conference (USENIX ATC 20)}, pages 219--233. USENIX Association, July 2020.
\newblock ISBN 978-1-939133-14-4.
\newblock URL \url{https://www.usenix.org/conference/atc20/presentation/keahey}.

\bibitem[Li et~al.(2025{\natexlab{a}})Li, Sun, Cheng, Zhu, and Chen]{li2025chain}
Xue Li, Yiyou Sun, Wei Cheng, Yinglun Zhu, and Haifeng Chen.
\newblock Chain-of-region: Visual language models need details for diagram analysis.
\newblock In \emph{The Thirteenth International Conference on Learning Representations}, 2025{\natexlab{a}}.

\bibitem[Li et~al.(2024)Li, Gao, Zhao, Wang, Sun, Lyu, Hawkins, Vasconcelos, Golan, Luo, et~al.]{li2024core}
Yijiang Li, Qingying Gao, Tianwei Zhao, Bingyang Wang, Haoran Sun, Haiyun Lyu, Robert~D Hawkins, Nuno Vasconcelos, Tal Golan, Dezhi Luo, et~al.
\newblock Core knowledge deficits in multi-modal language models.
\newblock \emph{arXiv preprint arXiv:2410.10855}, 2024.

\bibitem[Li et~al.(2025{\natexlab{b}})Li, Ma, Li, Li, Rong, Xu, Zhang, Zhao, and Huang]{li2025star}
Zongzhao Li, Zongyang Ma, Mingze Li, Songyou Li, Yu~Rong, Tingyang Xu, Ziqi Zhang, Deli Zhao, and Wenbing Huang.
\newblock Star-r1: Spatial transformation reasoning by reinforcing multimodal llms.
\newblock \emph{arXiv preprint arXiv:2505.15804}, 2025{\natexlab{b}}.

\bibitem[Lin et~al.(2021)Lin, Wu, Yang, Lee, and Ren]{lin-etal-2021-riddlesense}
Bill~Yuchen Lin, Ziyi Wu, Yichi Yang, Dong-Ho Lee, and Xiang Ren.
\newblock {R}iddle{S}ense: Reasoning about riddle questions featuring linguistic creativity and commonsense knowledge.
\newblock In Chengqing Zong, Fei Xia, Wenjie Li, and Roberto Navigli, editors, \emph{Findings of the Association for Computational Linguistics: ACL-IJCNLP 2021}, pages 1504--1515, Online, August 2021. Association for Computational Linguistics.
\newblock \doi{10.18653/v1/2021.findings-acl.131}.
\newblock URL \url{https://aclanthology.org/2021.findings-acl.131/}.

\bibitem[Liu et~al.(2024)Liu, Feng, Xue, Wang, Wu, Lu, Zhao, Deng, Zhang, Ruan, et~al.]{liu2024deepseek}
Aixin Liu, Bei Feng, Bing Xue, Bingxuan Wang, Bochao Wu, Chengda Lu, Chenggang Zhao, Chengqi Deng, Chenyu Zhang, Chong Ruan, et~al.
\newblock Deepseek-v3 technical report.
\newblock \emph{arXiv preprint arXiv:2412.19437}, 2024.

\bibitem[Liu et~al.(2023)Liu, Li, Wu, and Lee]{llava}
Haotian Liu, Chunyuan Li, Qingyang Wu, and Yong~Jae Lee.
\newblock Visual instruction tuning.
\newblock In A.~Oh, T.~Naumann, A.~Globerson, K.~Saenko, M.~Hardt, and S.~Levine, editors, \emph{Advances in Neural Information Processing Systems}, volume~36, pages 34892--34916. Curran Associates, Inc., 2023.
\newblock URL \url{https://proceedings.neurips.cc/paper_files/paper/2023/file/6dcf277ea32ce3288914faf369fe6de0-Paper-Conference.pdf}.

\bibitem[Liu et~al.(2020)Liu, Cui, Liu, Huang, Wang, and Zhang]{liu2020logiqa}
Jian Liu, Leyang Cui, Hanmeng Liu, Dandan Huang, Yile Wang, and Yue Zhang.
\newblock Logiqa: A challenge dataset for machine reading comprehension with logical reasoning.
\newblock \emph{arXiv preprint arXiv:2007.08124}, 2020.

\bibitem[Lu et~al.(2023)Lu, Bansal, Xia, Liu, Li, Hajishirzi, Cheng, Chang, Galley, and Gao]{lu2023mathvista}
Pan Lu, Hritik Bansal, Tony Xia, Jiacheng Liu, Chunyuan Li, Hannaneh Hajishirzi, Hao Cheng, Kai-Wei Chang, Michel Galley, and Jianfeng Gao.
\newblock Mathvista: Evaluating mathematical reasoning of foundation models in visual contexts.
\newblock \emph{arXiv preprint arXiv:2310.02255}, 2023.

\bibitem[Luo et~al.(2024)Luo, Zhang, Chen, Lin, Liu, Wu, Yang, Wang, Zeng, Gao, et~al.]{luo2024mmevol}
Run Luo, Haonan Zhang, Longze Chen, Ting-En Lin, Xiong Liu, Yuchuan Wu, Min Yang, Minzheng Wang, Pengpeng Zeng, Lianli Gao, et~al.
\newblock Mmevol: Empowering multimodal large language models with evol-instruct.
\newblock \emph{arXiv preprint arXiv:2409.05840}, 2024.

\bibitem[Meng et~al.(2025)Meng, Du, Liu, Zhou, Lu, Fu, Han, Shi, Wang, He, et~al.]{MMEureka}
Fanqing Meng, Lingxiao Du, Zongkai Liu, Zhixiang Zhou, Quanfeng Lu, Daocheng Fu, Tiancheng Han, Botian Shi, Wenhai Wang, Junjun He, et~al.
\newblock Mm-eureka: Exploring the frontiers of multimodal reasoning with rule-based reinforcement learning.
\newblock \emph{arXiv preprint arXiv:2503.07365}, 2025.

\bibitem[{Meta AI}(2025)]{meta2025llama4maverick}
{Meta AI}.
\newblock {The Llama 4 herd: The beginning of a new era of natively multimodal AI innovation}.
\newblock \url{https://ai.meta.com/blog/llama-4-multimodal-intelligence/}, April 2025.
\newblock Accessed: 2025-04-05.

\bibitem[OLMo et~al.(2025)OLMo, Walsh, Soldaini, Groeneveld, Lo, Arora, Bhagia, Gu, Huang, Jordan, Lambert, Schwenk, Tafjord, Anderson, Atkinson, Brahman, Clark, Dasigi, Dziri, Guerquin, Ivison, Koh, Liu, Malik, Merrill, Miranda, Morrison, Murray, Nam, Pyatkin, Rangapur, Schmitz, Skjonsberg, Wadden, Wilhelm, Wilson, Zettlemoyer, Farhadi, Smith, and Hajishirzi]{olmo20252olmo2furious}
Team OLMo, Pete Walsh, Luca Soldaini, Dirk Groeneveld, Kyle Lo, Shane Arora, Akshita Bhagia, Yuling Gu, Shengyi Huang, Matt Jordan, Nathan Lambert, Dustin Schwenk, Oyvind Tafjord, Taira Anderson, David Atkinson, Faeze Brahman, Christopher Clark, Pradeep Dasigi, Nouha Dziri, Michal Guerquin, Hamish Ivison, Pang~Wei Koh, Jiacheng Liu, Saumya Malik, William Merrill, Lester James~V. Miranda, Jacob Morrison, Tyler Murray, Crystal Nam, Valentina Pyatkin, Aman Rangapur, Michael Schmitz, Sam Skjonsberg, David Wadden, Christopher Wilhelm, Michael Wilson, Luke Zettlemoyer, Ali Farhadi, Noah~A. Smith, and Hannaneh Hajishirzi.
\newblock 2 olmo 2 furious, 2025.
\newblock URL \url{https://arxiv.org/abs/2501.00656}.

\bibitem[{OpenAI}(2025{\natexlab{a}})]{openai2025introducing4o}
{OpenAI}.
\newblock Introducing 4o image generation.
\newblock OpenAI Website, March 2025{\natexlab{a}}.
\newblock URL \url{https://openai.com/index/introducing-4o-image-generation/}.
\newblock Accessed: 2025-05-16.

\bibitem[{OpenAI}(2025{\natexlab{b}})]{openai2025o3o4mini}
{OpenAI}.
\newblock Openai o3 and o4-mini system card.
\newblock OpenAI Website, April 2025{\natexlab{b}}.
\newblock URL \url{https://openai.com/index/o3-o4-mini-system-card/}.
\newblock Accessed: 2025-05-16.

\bibitem[Qiao et~al.(2024)Qiao, Tan, Dong, Wu, Sun, Song, GongQue, Lei, Wei, Zhang, et~al.]{qiao2407we}
Runqi Qiao, Qiuna Tan, Guanting Dong, Minhui Wu, Chong Sun, Xiaoshuai Song, Zhuoma GongQue, Shanglin Lei, Zhe Wei, Miaoxuan Zhang, et~al.
\newblock We-math: Does your large multimodal model achieve human-like mathematical reasoning?
\newblock \emph{arXiv preprint arXiv:2407.01284}, 2024.

\bibitem[QwenTeam(2025)]{qwen2.5-VL}
QwenTeam.
\newblock Qwen2.5-vl, January 2025.
\newblock URL \url{https://qwenlm.github.io/blog/qwen2.5-vl/}.

\bibitem[Radford et~al.(2021)Radford, Kim, Hallacy, Ramesh, Goh, Agarwal, Sastry, Askell, Mishkin, Clark, et~al.]{radford2021learning}
Alec Radford, Jong~Wook Kim, Chris Hallacy, Aditya Ramesh, Gabriel Goh, Sandhini Agarwal, Girish Sastry, Amanda Askell, Pamela Mishkin, Jack Clark, et~al.
\newblock Learning transferable visual models from natural language supervision.
\newblock In \emph{International conference on machine learning}, pages 8748--8763. PmLR, 2021.

\bibitem[Ren et~al.(2025)Ren, Tertikas, Maiti, Han, Zhang, S{\"u}sstrunk, and Kokkinos]{ren2025vgrp}
Yufan Ren, Konstantinos Tertikas, Shalini Maiti, Junlin Han, Tong Zhang, Sabine S{\"u}sstrunk, and Filippos Kokkinos.
\newblock Vgrp-bench: Visual grid reasoning puzzle benchmark for large vision-language models.
\newblock \emph{arXiv preprint arXiv:2503.23064}, 2025.

\bibitem[Rozner et~al.(2021)Rozner, Potts, and Mahowald]{rozner2021decrypting}
Joshua Rozner, Christopher Potts, and Kyle Mahowald.
\newblock Decrypting cryptic crosswords: Semantically complex wordplay puzzles as a target for {NLP}.
\newblock In A.~Beygelzimer, Y.~Dauphin, P.~Liang, and J.~Wortman Vaughan, editors, \emph{Advances in Neural Information Processing Systems}, 2021.
\newblock URL \url{https://openreview.net/forum?id=136ihvjd0sJ}.

\bibitem[Schuster et~al.(2021)Schuster, Kalyan, Polozov, and Kalai]{schuster2021programming}
Tal Schuster, Ashwin Kalyan, Alex Polozov, and Adam~Tauman Kalai.
\newblock Programming puzzles.
\newblock In \emph{Thirty-fifth Conference on Neural Information Processing Systems Datasets and Benchmarks Track (Round 1)}, 2021.
\newblock URL \url{https://openreview.net/forum?id=fe_hCc4RBrg}.

\bibitem[Shao et~al.(2024)Shao, Wang, Zhu, Xu, Song, Bi, Zhang, Zhang, Li, Wu, et~al.]{shao2024deepseekmath}
Zhihong Shao, Peiyi Wang, Qihao Zhu, Runxin Xu, Junxiao Song, Xiao Bi, Haowei Zhang, Mingchuan Zhang, YK~Li, Y~Wu, et~al.
\newblock Deepseekmath: Pushing the limits of mathematical reasoning in open language models.
\newblock \emph{arXiv preprint arXiv:2402.03300}, 2024.

\bibitem[Sheng et~al.(2025)Sheng, Zhang, Ye, Wu, Zhang, Zhang, Peng, Lin, and Wu]{Sheng_2025}
Guangming Sheng, Chi Zhang, Zilingfeng Ye, Xibin Wu, Wang Zhang, Ru~Zhang, Yanghua Peng, Haibin Lin, and Chuan Wu.
\newblock Hybridflow: A flexible and efficient rlhf framework.
\newblock In \emph{Proceedings of the Twentieth European Conference on Computer Systems}, EuroSys ’25, page 1279–1297. ACM, March 2025.
\newblock \doi{10.1145/3689031.3696075}.
\newblock URL \url{http://dx.doi.org/10.1145/3689031.3696075}.

\bibitem[Song et~al.(2025)Song, Ou, Kong, Li, Neubig, and Yue]{song2025visualpuzzles}
Yueqi Song, Tianyue Ou, Yibo Kong, Zecheng Li, Graham Neubig, and Xiang Yue.
\newblock Visualpuzzles: Decoupling multimodal reasoning evaluation from domain knowledge.
\newblock \emph{arXiv preprint arXiv:2504.10342}, 2025.

\bibitem[Touvron et~al.(2023)Touvron, Martin, Stone, Albert, Almahairi, Babaei, Bashlykov, Batra, Bhargava, Bhosale, Bikel, Blecher, Ferrer, Chen, Cucurull, Esiobu, Fernandes, Fu, Fu, Fuller, Gao, Goswami, Goyal, Hartshorn, Hosseini, Hou, Inan, Kardas, Kerkez, Khabsa, Kloumann, Korenev, Koura, Lachaux, Lavril, Lee, Liskovich, Lu, Mao, Martinet, Mihaylov, Mishra, Molybog, Nie, Poulton, Reizenstein, Rungta, Saladi, Schelten, Silva, Smith, Subramanian, Tan, Tang, Taylor, Williams, Kuan, Xu, Yan, Zarov, Zhang, Fan, Kambadur, Narang, Rodriguez, Stojnic, Edunov, and Scialom]{touvron2023llama2openfoundation}
Hugo Touvron, Louis Martin, Kevin Stone, Peter Albert, Amjad Almahairi, Yasmine Babaei, Nikolay Bashlykov, Soumya Batra, Prajjwal Bhargava, Shruti Bhosale, Dan Bikel, Lukas Blecher, Cristian~Canton Ferrer, Moya Chen, Guillem Cucurull, David Esiobu, Jude Fernandes, Jeremy Fu, Wenyin Fu, Brian Fuller, Cynthia Gao, Vedanuj Goswami, Naman Goyal, Anthony Hartshorn, Saghar Hosseini, Rui Hou, Hakan Inan, Marcin Kardas, Viktor Kerkez, Madian Khabsa, Isabel Kloumann, Artem Korenev, Punit~Singh Koura, Marie-Anne Lachaux, Thibaut Lavril, Jenya Lee, Diana Liskovich, Yinghai Lu, Yuning Mao, Xavier Martinet, Todor Mihaylov, Pushkar Mishra, Igor Molybog, Yixin Nie, Andrew Poulton, Jeremy Reizenstein, Rashi Rungta, Kalyan Saladi, Alan Schelten, Ruan Silva, Eric~Michael Smith, Ranjan Subramanian, Xiaoqing~Ellen Tan, Binh Tang, Ross Taylor, Adina Williams, Jian~Xiang Kuan, Puxin Xu, Zheng Yan, Iliyan Zarov, Yuchen Zhang, Angela Fan, Melanie Kambadur, Sharan Narang, Aurelien Rodriguez, Robert Stojnic, Sergey Edunov, and Thomas
  Scialom.
\newblock Llama 2: Open foundation and fine-tuned chat models, 2023.
\newblock URL \url{https://arxiv.org/abs/2307.09288}.

\bibitem[Van~der Maaten and Hinton(2008)]{van2008visualizing}
Laurens Van~der Maaten and Geoffrey Hinton.
\newblock Visualizing data using t-sne.
\newblock \emph{Journal of machine learning research}, 9\penalty0 (11), 2008.

\bibitem[Wang et~al.(2025{\natexlab{a}})Wang, Li, Zhou, Leong, Zhao, Ye, Deng, Luo, and Vasconcelos]{wang2025vision}
Bingyang Wang, Yijiang Li, Qingyang Zhou, Hui~Yi Leong, Tianwei Zhao, Letian Ye, Hokin Deng, Dezhi Luo, and Nuno Vasconcelos.
\newblock Do vision language models infer human intention without visual perspective-taking? towards a scalable" one-image-probe-all" dataset.
\newblock In \emph{ICML 2025 Workshop on Assessing World Models}, 2025{\natexlab{a}}.

\bibitem[Wang et~al.(2025{\natexlab{b}})Wang, Zhou, Xu, Cheng, Zuo, Tian, Song, Lu, Hu, and Liu]{wang2025code}
Hanbin Wang, Xiaoxuan Zhou, Zhipeng Xu, Keyuan Cheng, Yuxin Zuo, Kai Tian, Jingwei Song, Junting Lu, Wenhui Hu, and Xueyang Liu.
\newblock Code-vision: Evaluating multimodal llms logic understanding and code generation capabilities.
\newblock \emph{arXiv preprint arXiv:2502.11829}, 2025{\natexlab{b}}.

\bibitem[Wang et~al.(2023)Wang, Kordi, Mishra, Liu, Smith, Khashabi, and Hajishirzi]{wang-etal-2023-self-instruct}
Yizhong Wang, Yeganeh Kordi, Swaroop Mishra, Alisa Liu, Noah~A. Smith, Daniel Khashabi, and Hannaneh Hajishirzi.
\newblock Self-instruct: Aligning language models with self-generated instructions.
\newblock In \emph{Proceedings of the 61st Annual Meeting of the Association for Computational Linguistics (Volume 1: Long Papers)}, pages 13484--13508, Toronto, Canada, 2023. Association for Computational Linguistics.

\bibitem[Wang et~al.(2004)Wang, Bovik, Sheikh, and Simoncelli]{wang2004image}
Zhou Wang, Alan~C Bovik, Hamid~R Sheikh, and Eero~P Simoncelli.
\newblock Image quality assessment: from error visibility to structural similarity.
\newblock \emph{IEEE transactions on image processing}, 13\penalty0 (4):\penalty0 600--612, 2004.

\bibitem[Wang et~al.(2025{\natexlab{c}})Wang, Zhu, Tang, Li, Xiong, Yu, and Blaschko]{wang2025jigsaw}
Zifu Wang, Junyi Zhu, Bo~Tang, Zhiyu Li, Feiyu Xiong, Jiaqian Yu, and Matthew~B. Blaschko.
\newblock Jigsaw-r1: A study of rule-based visual reinforcement learning with jigsaw puzzles.
\newblock \emph{Transactions on Machine Learning Research}, 2025{\natexlab{c}}.
\newblock URL \url{https://arxiv.org/abs/2505.23590}.

\bibitem[xAI(2025)]{xai2025grok3}
xAI.
\newblock Grok 3 beta — the age of reasoning agents.
\newblock \url{https://x.ai/blog/grok-3}, 2025.
\newblock Accessed: 2025-02-21.

\bibitem[Xu et~al.(2023)Xu, Sun, Zheng, Geng, Zhao, Feng, Tao, and Jiang]{xu2023wizardlm}
Can Xu, Qingfeng Sun, Kai Zheng, Xiubo Geng, Pu~Zhao, Jiazhan Feng, Chongyang Tao, and Daxin Jiang.
\newblock Wizardlm: Empowering large language models to follow complex instructions.
\newblock \emph{arXiv preprint arXiv:2304.12244}, 2023.

\bibitem[Xu et~al.(2025)Xu, Wang, Wang, Chen, Zhou, Yang, Lu, Li, Wang, Zhu, et~al.]{xu2025visulogic}
Weiye Xu, Jiahao Wang, Weiyun Wang, Zhe Chen, Wengang Zhou, Aijun Yang, Lewei Lu, Houqiang Li, Xiaohua Wang, Xizhou Zhu, et~al.
\newblock Visulogic: A benchmark for evaluating visual reasoning in multi-modal large language models.
\newblock \emph{arXiv preprint arXiv:2504.15279}, 2025.

\bibitem[Xu et~al.(2024)Xu, Jiang, Niu, Deng, Poovendran, Choi, and Lin]{xu2024magpie}
Zhangchen Xu, Fengqing Jiang, Luyao Niu, Yuntian Deng, Radha Poovendran, Yejin Choi, and Bill~Yuchen Lin.
\newblock Magpie: Alignment data synthesis from scratch by prompting aligned llms with nothing.
\newblock \emph{arXiv preprint arXiv:2406.08464}, 2024.

\bibitem[Yang et~al.(2025)Yang, He, Pan, Jiang, Deng, Yang, Lu, Yin, Rao, Zhu, et~al.]{yang2025r1}
Yi~Yang, Xiaoxuan He, Hongkun Pan, Xiyan Jiang, Yan Deng, Xingtao Yang, Haoyu Lu, Dacheng Yin, Fengyun Rao, Minfeng Zhu, et~al.
\newblock R1-onevision: Advancing generalized multimodal reasoning through cross-modal formalization.
\newblock \emph{arXiv preprint arXiv:2503.10615}, 2025.

\bibitem[Yao et~al.(2023)Yao, Yu, Zhao, Shafran, Griffiths, Cao, and Narasimhan]{yao2023tree}
Shunyu Yao, Dian Yu, Jeffrey Zhao, Izhak Shafran, Thomas~L. Griffiths, Yuan Cao, and Karthik~R Narasimhan.
\newblock Tree of thoughts: Deliberate problem solving with large language models.
\newblock In \emph{Thirty-seventh Conference on Neural Information Processing Systems}, 2023.
\newblock URL \url{https://openreview.net/forum?id=5Xc1ecxO1h}.

\bibitem[Ye et~al.(2023)Ye, Hu, Xu, Ye, Yan, Xu, Li, Tian, Qian, Zhang, Jin, He, Lin, and Huang]{ye-etal-2023-ureader}
Jiabo Ye, Anwen Hu, Haiyang Xu, Qinghao Ye, Ming Yan, Guohai Xu, Chenliang Li, Junfeng Tian, Qi~Qian, Ji~Zhang, Qin Jin, Liang He, Xin Lin, and Fei Huang.
\newblock {UR}eader: Universal {OCR}-free visually-situated language understanding with multimodal large language model.
\newblock In Houda Bouamor, Juan Pino, and Kalika Bali, editors, \emph{Findings of the Association for Computational Linguistics: EMNLP 2023}, pages 2841--2858, Singapore, December 2023. Association for Computational Linguistics.
\newblock \doi{10.18653/v1/2023.findings-emnlp.187}.
\newblock URL \url{https://aclanthology.org/2023.findings-emnlp.187/}.

\bibitem[Yue et~al.(2024)Yue, Zheng, Ni, Wang, Zhang, Tong, Sun, Yu, Zhang, Sun, et~al.]{yue2024mmmu}
Xiang Yue, Tianyu Zheng, Yuansheng Ni, Yubo Wang, Kai Zhang, Shengbang Tong, Yuxuan Sun, Botao Yu, Ge~Zhang, Huan Sun, et~al.
\newblock Mmmu-pro: A more robust multi-discipline multimodal understanding benchmark.
\newblock \emph{arXiv preprint arXiv:2409.02813}, 2024.

\bibitem[Zhai et~al.(2023)Zhai, Mustafa, Kolesnikov, and Beyer]{zhai2023sigmoid}
Xiaohua Zhai, Basil Mustafa, Alexander Kolesnikov, and Lucas Beyer.
\newblock Sigmoid loss for language image pre-training.
\newblock In \emph{Proceedings of the IEEE/CVF international conference on computer vision}, pages 11975--11986, 2023.

\bibitem[Zhang et~al.(2019)Zhang, Gao, Jia, Zhu, and Zhu]{zhang2019raven}
Chi Zhang, Feng Gao, Baoxiong Jia, Yixin Zhu, and Song-Chun Zhu.
\newblock Raven: A dataset for relational and analogical visual reasoning.
\newblock In \emph{Proceedings of the IEEE/CVF Conference on Computer Vision and Pattern Recognition (CVPR)}, pages 5317--5327, 2019.

\bibitem[Zhang et~al.(2024{\natexlab{a}})Zhang, Jiang, Zhang, Lin, Guo, Qiu, Zhou, Lu, Chang, Qiao, et~al.]{zhang2024mathverse}
Renrui Zhang, Dongzhi Jiang, Yichi Zhang, Haokun Lin, Ziyu Guo, Pengshuo Qiu, Aojun Zhou, Pan Lu, Kai-Wei Chang, Yu~Qiao, et~al.
\newblock Mathverse: Does your multi-modal llm truly see the diagrams in visual math problems?
\newblock In \emph{European Conference on Computer Vision}, pages 169--186. Springer, 2024{\natexlab{a}}.

\bibitem[Zhang et~al.(2025{\natexlab{a}})Zhang, Feng, Luo, Li, Yue, Zheng, and Lu]{zhang2025unified}
Wanpeng Zhang, Yicheng Feng, Hao Luo, Yijiang Li, Zihao Yue, Sipeng Zheng, and Zongqing Lu.
\newblock Unified multimodal understanding via byte-pair visual encoding.
\newblock \emph{arXiv preprint arXiv:2506.23639}, 2025{\natexlab{a}}.

\bibitem[Zhang et~al.(2024{\natexlab{b}})Zhang, Cheng, He, Wang, Shen, Tan, Hou, He, Ma, Lu, et~al.]{zhang2024multimodal}
Wenqi Zhang, Zhenglin Cheng, Yuanyu He, Mengna Wang, Yongliang Shen, Zeqi Tan, Guiyang Hou, Mingqian He, Yanna Ma, Weiming Lu, et~al.
\newblock Multimodal self-instruct: Synthetic abstract image and visual reasoning instruction using language model.
\newblock \emph{arXiv preprint arXiv:2407.07053}, 2024{\natexlab{b}}.

\bibitem[Zhang et~al.(2025{\natexlab{b}})Zhang, Yu, Tian, Fu, Li, Zeng, Xie, Shi, Zhang, Wu, et~al.]{zhang2025mm}
Yi-Fan Zhang, Tao Yu, Haochen Tian, Chaoyou Fu, Peiyan Li, Jianshu Zeng, Wulin Xie, Yang Shi, Huanyu Zhang, Junkang Wu, et~al.
\newblock Mm-rlhf: The next step forward in multimodal llm alignment.
\newblock \emph{arXiv preprint arXiv:2502.10391}, 2025{\natexlab{b}}.

\bibitem[Zhang et~al.(2025{\natexlab{c}})Zhang, Chen, Zhang, Sun, Tian, Jia, Li, Liu, Min, and Zhai]{zhang2025puzzlebench}
Zeyu Zhang, Zijian Chen, Zicheng Zhang, Yuze Sun, Yuan Tian, Ziheng Jia, Chunyi Li, Xiaohong Liu, Xiongkuo Min, and Guangtao Zhai.
\newblock Puzzlebench: A fully dynamic evaluation framework for large multimodal models on puzzle solving.
\newblock \emph{arXiv preprint arXiv:2504.10885}, 2025{\natexlab{c}}.

\bibitem[Zheng et~al.(2024)Zheng, Zhang, Zhang, Ye, Luo, Feng, and Ma]{zheng2024llamafactory}
Yaowei Zheng, Richong Zhang, Junhao Zhang, Yanhan Ye, Zheyan Luo, Zhangchi Feng, and Yongqiang Ma.
\newblock Llamafactory: Unified efficient fine-tuning of 100+ language models.
\newblock In \emph{Proceedings of the 62nd Annual Meeting of the Association for Computational Linguistics (Volume 3: System Demonstrations)}, Bangkok, Thailand, 2024. Association for Computational Linguistics.
\newblock URL \url{http://arxiv.org/abs/2403.13372}.

\bibitem[Zheng et~al.(2025)Zheng, Lu, Wang, Feng, Kuang, and Xiong]{zheng2025easyr1}
Yaowei Zheng, Junting Lu, Shenzhi Wang, Zhangchi Feng, Dongdong Kuang, and Yuwen Xiong.
\newblock Easyr1: An efficient, scalable, multi-modality rl training framework, 2025.

\bibitem[Zhou et~al.(2025)Zhou, Qiu, Chen, Wang, Yang, Xu, and Qiu]{zhou2025reinforced}
Guanghao Zhou, Panjia Qiu, Cen Chen, Jie Wang, Zheming Yang, Jian Xu, and Minghui Qiu.
\newblock Reinforced mllm: A survey on rl-based reasoning in multimodal large language models.
\newblock \emph{arXiv preprint arXiv:2504.21277}, 2025.

\end{thebibliography}
